%% file: iclr2025_conference.tex
\documentclass{article} 
\usepackage{iclr2025_conference,times}

\input{math_commands.tex}

\usepackage{amsmath}
\usepackage{hyperref}
\usepackage{url}
\usepackage{multirow}
\usepackage{subfigure}
\usepackage{bbding}
\usepackage[ruled,vlined]{algorithm2e}
\usepackage{wrapfig}
\usepackage{array}

\usepackage{adjustbox}
\newcolumntype{P}[1]{>{\centering\arraybackslash}p{#1}}

\title{Revisiting the Othello World Model Hypothesis}
\iclrfinalcopy

\author{Yifei Yuan, Anders Søgaard \\
University of Copenhagen, Denmark\\
\texttt{\{yiya, soegaard\}@di.ku.dk} \\
}

\begin{document}

\maketitle

\begin{abstract}
\citet{li2023emergent} used the Othello board game as a test case for the ability of GPT-2 to induce world models, and were followed up by \citet{nanda-etal-2023-emergent}. We briefly discuss the original experiments, expanding them to include more language models with more comprehensive probing. Specifically, we analyze sequences of Othello board states and train the model to predict the next move based on previous moves. We evaluate seven language models (GPT-2, T5, Bart, Flan-T5, Mistral, LLaMA-2, and Qwen2.5) on the Othello task and conclude that these models not only learn to play Othello, but also induce the Othello board layout. We find that all models achieve up to 99\% accuracy in \textit{unsupervised} grounding and exhibit high similarity in the board features they learned. This provides considerably stronger evidence for the Othello World Model Hypothesis than previous works. 
\end{abstract}

\section{Introduction}

\citet{li2023emergent} used the Othello board game to probe the ability of LLMs to induce world models. Their network had a 60-word input vocabulary, corresponding to the 64 tiles of an Othello board, except for the four that are already filled at the start. They trained the network on two datasets: one on about 140,000 real Othello games and another on millions of synthetic games. They then trained 64 independent non-linear probes (two-layer MLP classifiers) to classify each of the 64 tiles into three states: black, blank, and white, using internal representations from Othello-GPT as input. 
The error rates of these non-linear probes dropped from 26.2\% on a randomly-initialized model to only 1.7\% on a trained model, while linear probes performed close to random. \citet{li2023emergent} saw this as evidence that LLMs can induce (non-linear) world models, at least for Othello board games, supporting the Othello World Model Hypothesis --  -- the hypothesis that LLMs trained on Othello move sequences can induce a relevant world model, including the Othello board layout.


\citet{nanda-etal-2023-emergent} did a follow-up study in which they found that linear probes also work if trained slightly differently. Instead of focusing on tile color, they probed the board state relative to the current player at each timestep, using labels such as MINE, YOURS, and EMPTY. This reduced the error rate of the probes to less than 10\%. They speculated that world knowledge is often linearly represented in language models, 
since `matrix
multiplication can easily extract a different subset
of linear features for each neuron.'  

Now, training a probe as a research methodology comes with several weaknesses, including: a) probing classifiers can be prone to spurious correlations \citep{barrett-etal-2019-adversarial}. b) They do not tell us how information is arranged globally in LLMs.\footnote{\citet{li2023emergent} tried to compensate for this by using PCA to plot the probing classifiers in three dimensions. The PCA plots suggest that the induced global structure is meaningful, but the probing paradigm cannot quantify its meaningfulness.} c) They therefore only detect a subset of the interesting properties of world models, e.g., excluding the spatial relations that would enable analogical reasoning \citep{NIPS2013_9aa42b31}. 

\begin{wrapfigure}{r}{7cm}
    \centering
    \includegraphics[width=0.95\linewidth]{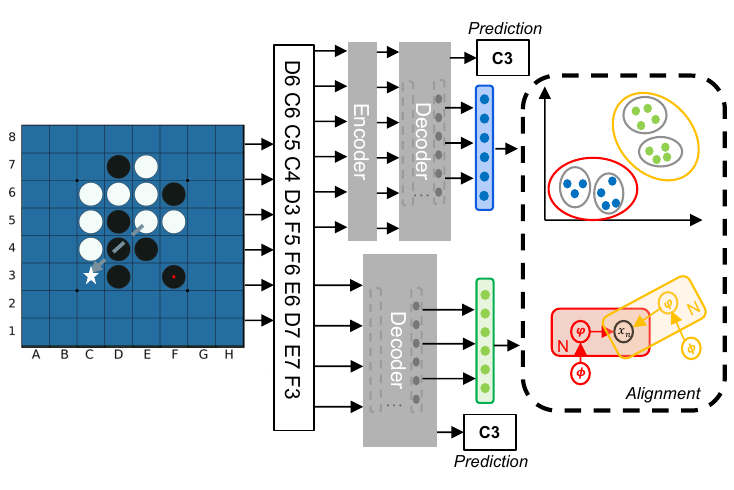}
    \caption{Experimental protocol. We re-train the Transformer-based models to predict the next move in Othello and see whether the board game layout is induced (up to isomorphism). }
    \label{fig:protocol}
\end{wrapfigure}

\paragraph{Contributions} We therefore revisit the Othello World Model Hypothesis, reevaluating it using a methodology that does not suffer from weaknesses a)--c) (see Figure \ref{fig:protocol}), in order to reassess the ability of LLMs to induce world models. If our results are positive, they will significantly strengthen the case for the hypothesis that LLMs induce world models; if not, they will suggest that the evidence cited in \citet{li2023emergent} and \citet{nanda-etal-2023-emergent} was perhaps a (spurious) effect of the probing paradigm itself. 
We begin by re-modeling Othello using a range of model sizes (GPT-2, BART, T5, Flan-T5, LLaMA-2, Mistral, Qwen2.5), as prior research has predominantly focused on smaller models like GPT-2. We retrain these models using game data of varying scales from the two datasets presented by \citet{li2023emergent}. Our analysis extends beyond previous studies by considering both pretrained and non-pretrained models (based on upstream language tasks), evaluating two-hop generation capabilities, and comparing models of varying sizes. To assess whether these models capture similar underlying game strategies, state representations, or other key aspects, despite differences in architecture and size, we employ representation alignment tools inspired by the literature on cross-lingual word embeddings \citep{4264a46fd9e846e4a704b2d13002e521}.
Finally, we visualize these results through latent move projections, enabling a deeper analysis of the internal mechanisms of models trained on the Othello game. Through these probing methods, we show that the language models -- exhibit solid one-hop performance when trained on large amount of game sequence moves. We find that in some cases, all models can achieve up to 99\% accuracy in unsupervised grounding, which means that absent any cross-modal supervision, a model {\em trained to play} Othello can identify the right positions on a board. More importantly, the alignment similarity score of the board features learned by these models is surprisingly high. Additionally, the latent move projection demonstrates that the models can learn the spatial structure of the chessboard. This provides a counter-example to previous claims that mono-modal models cannot solve visual question answering problems \citep{bender-koller-2020-climbing} -- or, more generally, symbol grounding problems \citep{Harnad1990-HARTSG}. Beyond that, these results are significantly stronger than those in \citet{li2023emergent,nanda-etal-2023-emergent} and, in our view, provide more direct evidence of the Othello World Model Hypothesis\footnote{Detailed definition see Appendix \ref{worldmodel}.}. 
\section{Related Work}

\paragraph{Past work on Othello}
Most past works on Othello~\citep{Chang2018TheBW, Ree2013ReinforcementLI} use reinforcement learning to search for moves. The first attempt to model Othello with deep neural networks dates back to 2018~\citep{8276588}, focusing on using CNNs to train a strong player. Based on it, ~\citet{noever2022word} focus on designing an effective Othello player with LLMs. Motivated by~\citet{Toshniwal2021LearningCB}, \citet{li2023emergent} shift the focus to treating the game as a diagnostic tool for inducing world models from text. Following this, \citet{nanda-etal-2023-emergent} provide evidence of a closely related linear representation of the board and propose a simple yet powerful way to interpret the model’s internal state. \citet{takizawa2024othello} recently presents a provably optimal strategy for playing Othello, delving into the complexity of these strategies and paving the way for future research to explore whether LLMs adopt similar approaches. \citet{Hua2024mOthelloWD} adopt the idea of Othello sequence generation and introduce a multilingual Othello task to aid in cross-lingual representation alignment.
\paragraph{World models} 
The success of language models in NLP tasks, to many, seems to turn on their ability to simulate, predict, and reason about dynamic environments as portrayed in  text~\citep{Hao2023ReasoningWL,Huh2024ThePR,Patel2022MappingLM,Xiang2023LanguageMM}. The seminal work of \citet{Li2021ImplicitRO} presents an example of fine-tuning LLMs on synthetic NLP tasks to find evidence that world states are weakly encoded in their activations. 
\citet{Wang2024CanLM} evaluate how well LLMs can serve as text-based world simulators with a benchmark. Inspired by Othello-GPT, research have explored more detailed probing~\citep{Yun2023EmergenceOA,Hazineh2023LinearLW} and more complex scenarios to assess the ability of LLMs to understand board states, including for games like chess, checker and maze navigation~\citep{Karvonen2024EmergentWM,Joshi2024CheckersGPTLW,Ivanitskiy2023StructuredWR}. Our work aims to revisit the Othello World Hypothesis using novel probing methods across a number of different LLMs.

\section{Modeling Othello with LLMs}
\label{section:modeling}
Following previous works~\citep{8276588,li2023emergent,nanda-etal-2023-emergent}, we formulate the problem of playing the board game as a sequence generation problem. Specifically, we fine-tune generative pretrained models in an autoregressive manner to predict the next move given the current Othello board state. We then evaluate whether the predicted move is legal or not. Each game is a sequence, with each move represented as a token, and in each round, we predict the next move. Our vocabulary consists of 60 words, each representing one of the 60 playable tiles where players place discs, excluding the 4 center tiles, which are already occupied at the start of the game. See Figure~\ref{fig:protocol} for an example move. Our modeling of Othello, in brief, can be represented as:
\begin{equation}
    p_{\theta}(X_{i+k}|X_{<i}) = \prod_{m=0}^{k}{p_{\theta}(X_{i+m}|X_{<i})}=\prod_{m=0}^{k}\mathit{softmax}(f_{i+m}(x_1,x_2,...,x_{i+m-1}))
\end{equation}
where $x_1, x_2, ..., x_{i-1}$ represent history moves, $X_{i+k}$ represents the sequence after $k$ generation steps.
During inference, we input the previously generated game moves $X_{<i}$ at step $i$ 
into the model and prompt it to generate the next steps. Unlike previous works, we not only prompt the model to generate the next move ($k=1$) but also introduce a new test where the model generates two consecutive moves ($k=2$), for it prompts models to simulate high-level reasoning, revealing how well LLMs understand game strategy in a zero-shot manner.

\subsection{Experimental Setup}
\input{tables/mainexp}
We use two datasets in our experiments, \textbf{CHAMPIONSHIP} and \textbf{SYNTHETIC}. Both of them were collected by \citet{li2023emergent}. \textbf{CHAMPIONSHIP} comes from real online Othello gaming sources, whereas \textbf{SYNTHETIC} is artificially generated according to the rules of Othello game play. Detailed statistics see Appendix \ref{datasetstatistics}. We use the last 20,000 games from each dataset for testing and validation (10,000 games each). Following \citet{li2023emergent}, we report the top-1 error rate of the generated move. That if a generated move is not legal, we count it as an error.  Specifically, we extend the 1-hop step generation setting in~\cite{li2023emergent} and investigate 2-hop move generation for investigating the model’s capability to anticipate more strategic, long-term planning in Othello.  This involves verifying whether the top-1 prediction is legal when the model is prompted to generate one and two moves at a time. We present the average error rate across all game sequences.
We implement all of the baselines under the Pytorch framework and the HuggingFace model repository. We conduct all of our experiments using 8 A100 GPUs. We use all the default parameters in the repository when fine-tuning.
\input{tables/mainexp1}

We perform our experiments using several existing baselines, with both Encoder-Decoder or Decoder-only structures. We first adopt some popular PLMs such as GPT-2~\citep{Radford2019LanguageMA}, T5~\citep{Raffel2019ExploringTL}, and Bart~\citep{Lewis2019BARTDS}. We then adopt several LLMs to see the their performance on this task, including Flan-T5~\citep{Chung2022ScalingIL}, LlaMa-2~\citep{Touvron2023Llama2O},  Mistral~\citep{Jiang2023Mistral7}, and Qwen2.5~\citep{Hui2024Qwen25CoderTR}. Details see Appendix \ref{sec:appendix2}.


\subsection{Evaluation of LLM Performance in Othello Move Generation}
\label{section:evaluation}
\begin{figure}[t]
\centering
    \subfigure[\textbf{CHAMPIONSHIP}]
    {
        \includegraphics[width=0.4\textwidth]{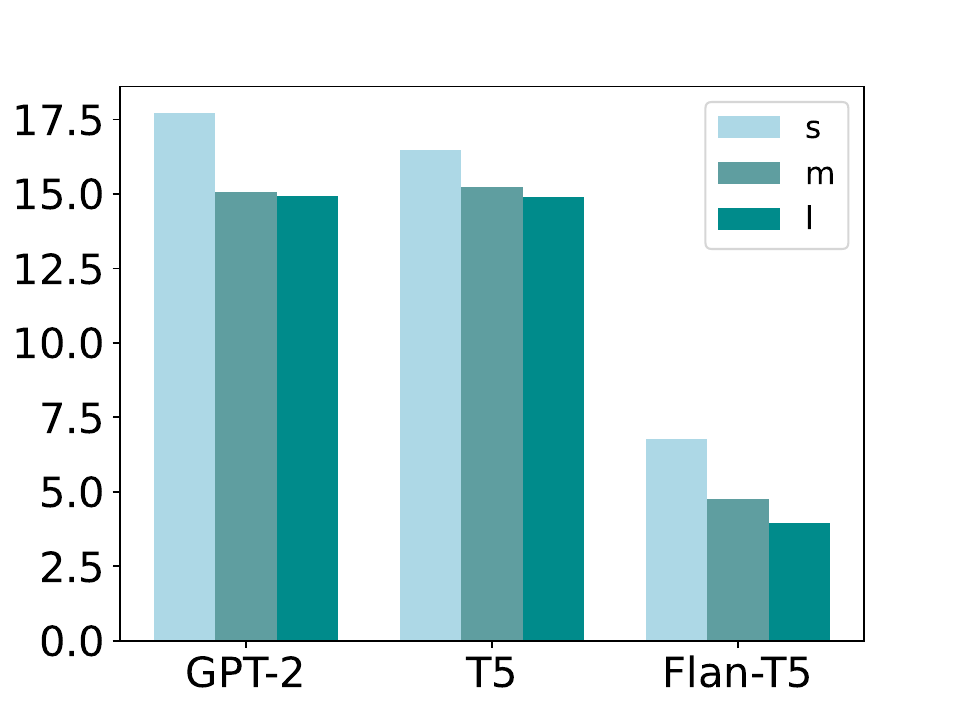}
        \label{fig:second_sub}
    }
    \subfigure[\textbf{SYNTHETIC}]
    {
        \includegraphics[width=0.4\textwidth]{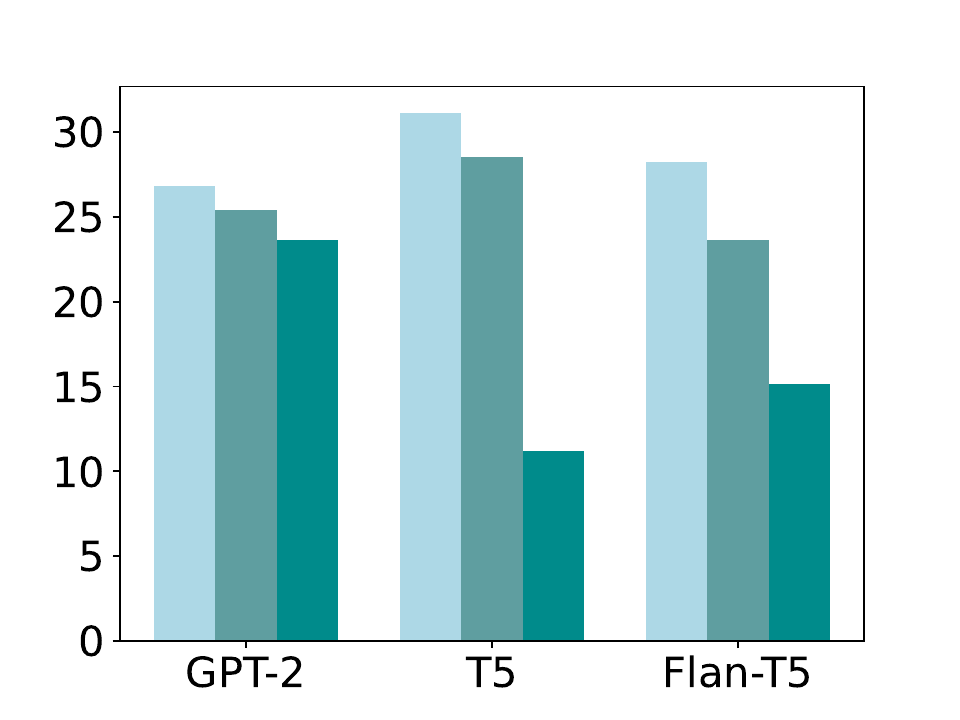}
        \label{fig:third_sub}
    }
    \caption{Othello 1-hop generation error rate under different model sizes. All models are non-pretrained versions fine-tuned with 20k game sequences.}
    \label{modelsizeanalysis}
\end{figure}
We perform experiments using various methods and present the results in Tables \ref{exp0} and \ref{exp1}. From our observations, several key findings emerge. Firstly, there is \textbf{no clear winner} between models with an Encoder-Decoder architecture, such as T5 or Flan-T5, and those with a Decoder-only architecture, such as GPT-2, LLaMA-2, or Qwen2.5 in terms of performance on this task. This indicates that the architectural differences between these models do not significantly impact their ability to generate Othello game steps. However, one consistent trend is the positive correlation between the amount of training data and overall model performance. As we increase the scale of the training data, all models tend to improve, underscoring the importance of data availability in mastering complex tasks like Othello move generation. In comparison to smaller language models, LLMs such as Mistral and Flan-T5 demonstrate clear superiority in this task, suggesting that \textbf{model size and capacity} are critical factors in understanding Othello game step generation. Larger models are better equipped to capture the intricate patterns and strategies within the game, likely due to their increased representational capacity. Interestingly, we also find that pretrained language knowledge, while generally beneficial for a wide range of natural language tasks, sometimes \textbf{negatively impacts} a model's ability to understand and generate game steps. Specifically, the pretrained versions of many models perform worse than their non-pretrained counterparts in this task, which could indicate that knowledge learned from upstream language tasks introduces biases or distracts from learning the specific structure and rules of Othello. Furthermore, while fine-tuning models on a large amount of data leads to reasonable performance in generating a single step (1-hop), generating more than one step consecutively remains a significant challenge. Even with large-scale data, models struggle to accurately predict two or more consecutive moves. This shows the potential limitation of the 1-hop evaluation since while it mostly focuses on the immediate next move based on the current board state, it inherently overlooks the deeper decision-making process required for gameplay strategies.

\subsection{Impact of Model Size on Othello Move Generation}
To further explore the impact of model size on the ability to model Othello moves, we analyze the performance of various models across different size configurations, as depicted in Figure \ref{modelsizeanalysis}. For each model, we evaluate performance in small, medium, and large size versions, allowing us to compare how scaling up model capacity affects accuracy in generating game moves. The results show a clear trend: \textbf{as model size increases, the error rate consistently decreases across both datasets}. This trend is particularly pronounced in the SYNTHETIC dataset, where larger models achieve significantly lower error rates compared to their smaller counterparts. The stronger improvement in the SYNTHETIC dataset suggests that larger models are better at capturing the structured patterns present in the synthetic data, likely due to their enhanced capacity for learning complex representations and generalizing across more varied scenarios. These findings highlight the importance of model scaling, showing that increasing the model size can lead to substantial performance gains in Othello move generation, especially in environments where the data is highly structured or synthetic in nature. Furthermore, the results emphasize that larger models are not just marginally better, but often significantly outperform smaller models, reinforcing the need to consider model capacity as a critical factor when tackling tasks that require a deep understanding of game strategies and sequential decision-making processes. 

\begin{figure}[t]
\centering
    \subfigure[\textbf{Non-pretrained}]
    {
        \includegraphics[width=0.43\textwidth,height=42mm]{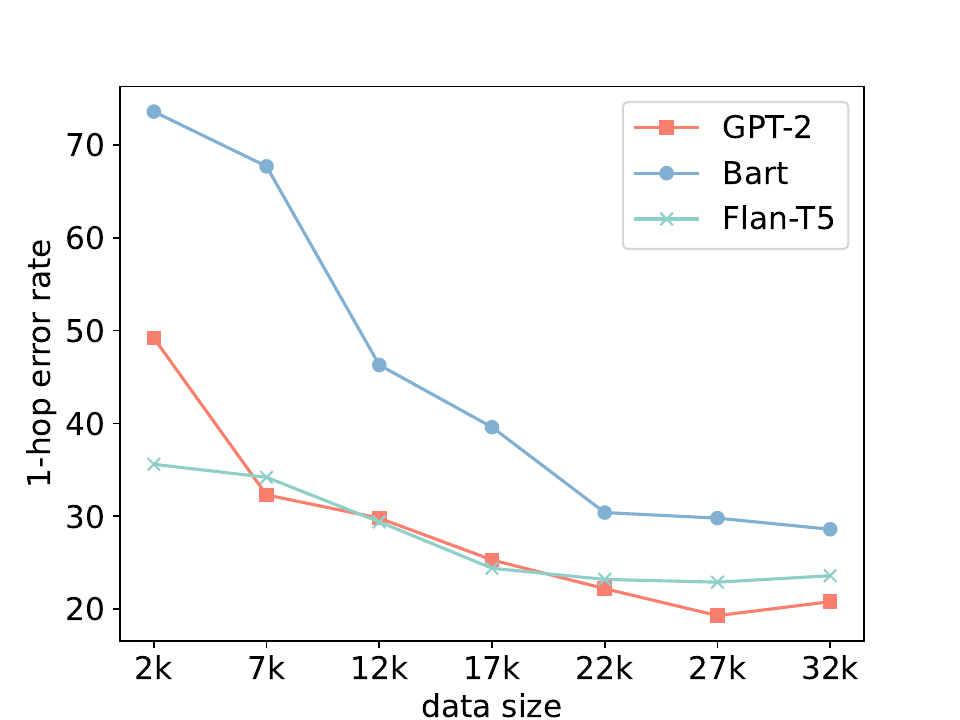}
        \label{fig:second_sub}
    }
    \subfigure[\textbf{Pretrained}]
    {
        \includegraphics[width=0.43\textwidth,height=42mm]{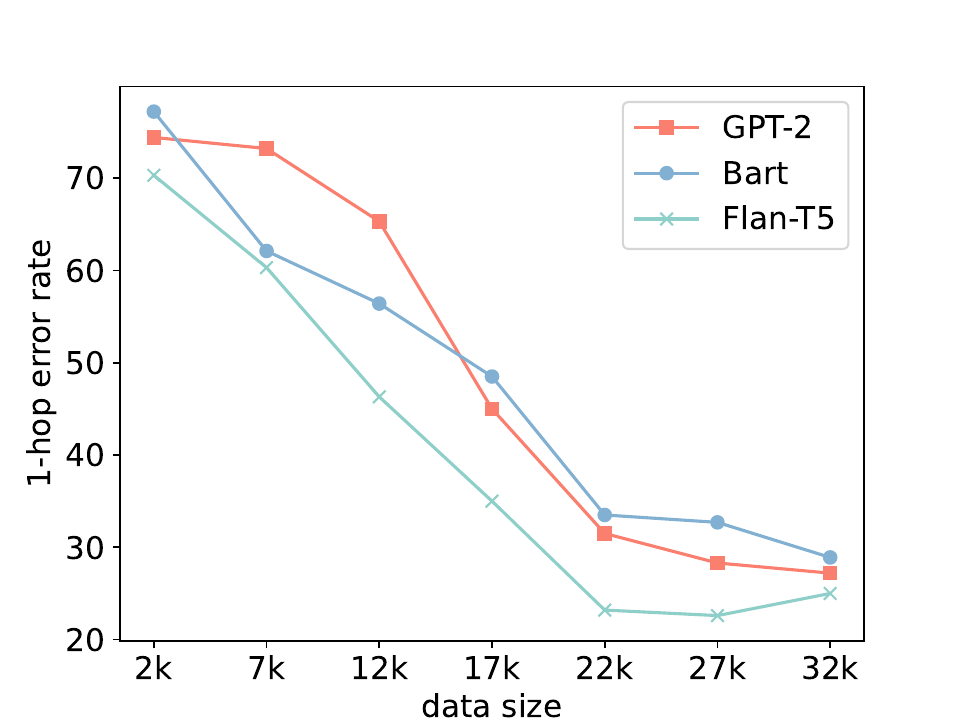}
        \label{fig:third_sub}
    }
    \caption{Analysis of 1-hop error rates on the SYNTHETIC dataset with varying data scales.}
    \label{sizeanalysis}
\end{figure}


\subsection{Impact of Data Size on Othello Move Generation}
In Table \ref{exp1}, we observe a sharp decrease in model error rates as the dataset size increases from 2k to 20k. To investigate this further, we conduct an analysis by gradually enlarging the SYNTHETIC dataset from 2k to 32k. According to Figure \ref{sizeanalysis}, the performance of all models improves gradually as the dataset size increases. Interestingly, non-pretrained models exhibit a faster reduction in error rates within the 2k to 12k data size range, with diminishing improvements beyond that point compared to pretrained models. In contrast, pretrained models take longer to achieve comparable performance, highlighting their slower adaptation to the task. This suggests that non-pretrained models are better suited for quickly learning game rules and adapting to fundamental patterns in the data. Furthermore, it indicates that the prior natural language knowledge embedded in pretrained models does not significantly enhance their understanding of the game. This observation aligns with our findings discussed in Section \ref{section:evaluation}, where we also observed the limited impact of pretrained knowledge in tasks requiring specialized domain adaptation.

\section{Othello Representation Alignment Across Language Models}
Drawing inspiration from the literature on cross-lingual word embeddings, we perform Othello representation alignment across different models to compare how each model, despite differences in architecture and size, internalizes and represents game strategies and states. This helps us assess whether the learned representations in Section \ref{section:modeling} are consistent across models and whether they capture similar underlying patterns essential for accurate Othello move generation.

\subsection{Alignment Method}
To validate the Othello World Model Hypothesis, we directly evaluate the internal representation of the Othello board in language models. Using the representations from different models, denoted as $F_1$, $F_2$ from the same input sequence $X_{<i}$, we perform mapping training under both \textit{supervised} and \textit{unsupervised} scenarios\footnote{Both of the algorithms are implemented using MUSE, a library designed for multilingual embedding alignment (\url{https://github.com/facebookresearch/MUSE}).}. A linear mapping $W$ is learned to map $F_1$ and $F_2$ into the same space:
\begin{equation}
    W^* = \mathop{\arg\min}_{W\in M_i(\mathbb R)}||WF_1-F_2||
\end{equation}
where $F_1, F_2 \in \mathbb R^{i\times h}$ are representations from the final hidden Decoder layer in different language models trained for Othello generation. 
$M_i(\mathbb R)$ is the space of $i \times i$ matrices of real numbers.


\textbf{Supervised training.} We consider the internal representations of different models within different source and target spaces. For supervised training (see Algorithm \ref{algo:supervised})\footnote{More details (e.g. BuildDict() of Algorithms \ref{algo:supervised}, \ref{algo:unsupervised}) see \cite{conneau2017word}.}, we use the pairwise data to learn a mapping from the source to the target space using iterative Procrustes alignment~\citep{gower2004procrustes}. We use representations from two models as training pairs. Specifically, the representations of the $i$th step within the same game from the two models are considered a pair, denoted as $h_{\theta1}(X_{<i})$ and $h_{\theta2}(X_{<i})$, respectively. In our experiment, we randomly select 1,000 game sequences from the validation set as training pairs. 
\begin{algorithm}[!ht]
\caption{Supervised Training for Othello Representation Alignment}
\small
\label{algo:supervised}
\DontPrintSemicolon
\SetKwInOut{Input}{Inputs}
\SetKwInOut{Output}{Output}
\SetInd{0.5em}{0.5em}
\Input{
\par
\begin{tabular}{l l}
$h_{\theta1}(\cdot)$, $h_{\theta2}(\cdot)$ & representations from the final hidden layer of Decoder in two models: $\Theta_1$, $\Theta_2$ \\
$X_{<i}=\{x_1,...,x_{i-1}\}$ & input game sequence at time step $i$ \\
$r$ & number of refinement iterations \\
\end{tabular}
}
\Output{
\par
\begin{tabular}{l l}
$s$ & Similarity score of the aligned feature learned from the two models
\end{tabular}
}

$ F_1 \gets h_{\theta1}(X_{<i}) $, 
$ F_2 \gets h_{\theta2}(X_{<i}) $ \\
\For{$i = 1$ to $r$}{ 
    \If {$i!=1$}{
    $F_1 \gets \operatorname{BuildDic} (F_1)$,  $F_2 \gets \operatorname{BuildDic} (F_2)$ \tcp*{build a dictionary from aligned embeddings containing best aligned pairs}} 
    $W \gets \operatorname{Procrustes} (F_1,F_2)$ \\
    $F_1 \gets WF_1$
}
$s \gets \operatorname{CosSim}(F_1, F_2)$

\end{algorithm}

\textbf{Unsupervised training.} For unsupervised training, without any parallel data or anchor points, following ~\cite{conneau2017word}, we learn the mapping through a combination of adversarial training and iterative Procrustes refinement~\citep{lample2017unsupervised} (see Algorithm \ref{algo:unsupervised}). The process involves first learning an initial proxy of the mapping $W$ using an adversarial criterion. Where an additional Discriminator model is trained to identify the origin of
an embedding, yet the target mapping $W$ aims at preventing the discriminator from doing so. Then, the mapping $W$ is further refined via
Procrustes using the same strategy in supervised training.  We then report the average cosine similarity of the aligned features on the test set. 
\begin{algorithm}[!ht]
\caption{Unsupervised Training for Othello Representation Alignment}
\small
\label{algo:unsupervised}
\DontPrintSemicolon
\SetKwInOut{Input}{Inputs}
\SetKwInOut{Output}{Output}
\SetInd{0.5em}{0.5em}
\Input{
\par
\begin{tabular}{l l}
$h_{\theta1}(\cdot)$, $h_{\theta2}(\cdot)$ & representations from the final hidden layer of Decoder in two models: $\Theta_1$, $\Theta_2$ \\
$X_{<i}=\{x_1,...,x_{i-1}\}$ & input game sequence at time step $i$ \\
$k, r$ & number of adversarial training iterations, number of refinement iterations \\
\end{tabular}
}
\Output{
\par
\begin{tabular}{l l}
$s$ & Similarity score of the aligned feature learned from the two models
\end{tabular}
}

$ F_1 \gets h_{\theta1}(X_{<i}) $, 
$ F_2 \gets h_{\theta2}(X_{<i}) $ \\
$\operatorname{RandomInitialize}(W)$ \\
\For{$i=1$ to $k$}{
$\mathcal D \gets \operatorname{TrainDiscriminator(\mathcal W,\mathcal D,F_1,F_2) }$ \tcp*{train the discriminator $\mathcal D$} 
$W \gets \operatorname{FoolDiscriminator(\mathcal W,\mathcal D,F_1,F_2)}$
\tcp*{train $W$ to fool the discriminator}
}
$ F_1 \gets WF_1$ \\
\For{$i=1$ to $r$ \tcp*{refine $W$}}{
$F_1 \gets \operatorname{BuildDic} (F_1)$,  $F_2 \gets \operatorname{BuildDic} (F_2)$ \\
$W \gets \operatorname{Procrustes} (F_1,F_2)$ \\
$F_1 \gets WF_1$
}
$s \gets CosSim(F_1, F_2)$

\end{algorithm}


\subsection{Mapping Result}
\label{representation_ali}
We probe different models by aligning their representations into one joint vector space. We report the cosine similarity of the aligned features score under both supervised~\citep{conneau2017word} and unsupervised~\citep{lample2017unsupervised} settings in Table \ref{exp2}\footnote{We use the non-pretrained version based on 20k  training data for all models.}. 
\input{tables/align}

From the results, we observe consistently high similarity scores across different language models, indicating that despite architectural differences, these models capture similar underlying representations when tasked with the Othello game. For instance, the SYNTHETIC supervised similarity score between GPT-2 (a Decoder-only model) and Bart (an Encoder-Decoder model) reaches an impressive 93.1\%. This suggests that, although these models process information differently due to their structural variances, they still converge on shared knowledge and representations when learning to model the Othello task. Such a high similarity score points to the possibility that both model types learn similar strategic patterns and rules intrinsic to the game, reinforcing the idea that fundamental aspects of the Othello task are captured across architectures. 
\subsection{PCA Visualization}
\begin{wrapfigure}{r}{0.5\linewidth}
    \centering
    \includegraphics[width=0.9\linewidth]{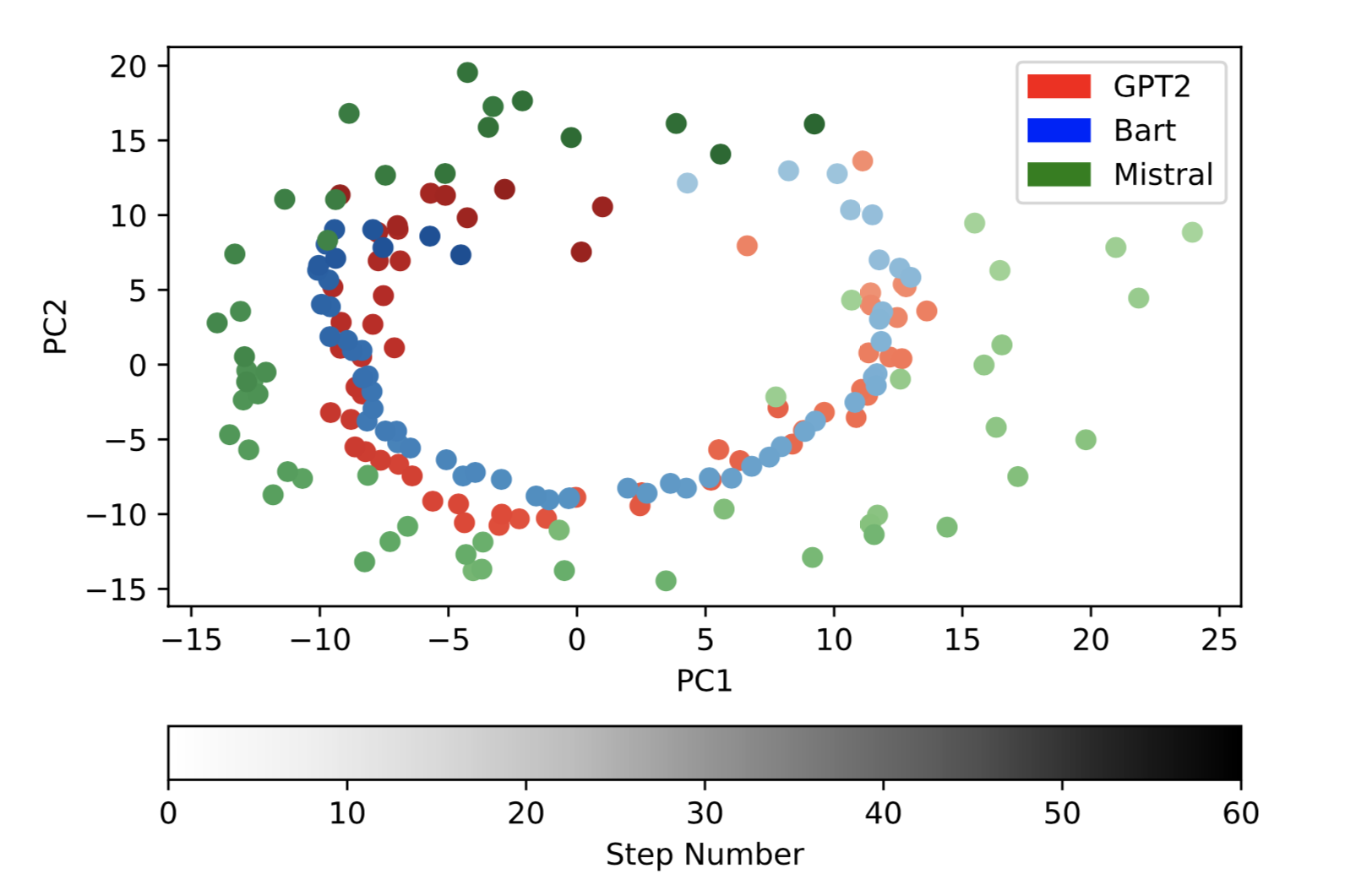}
    \caption{PCA visualization of the 60 steps from various models within one game.}
    \label{fig:alignment}
\end{wrapfigure}

In order to vividly show such alignment, we also demonstrate the dimension-reduced PCA  coordinate of 60 step features $h_\theta(X)$ within one entire random game in Figure \ref{fig:alignment}.  We also observe highly similar step representations across different models. This suggests that these models are learning comparable internal representations of the game states, indicating that the models are aligned in how they interpret the sequential nature of Othello.
Even though they may be built differently (e.g., Decoder-only versus Encoder-Decoder), the core representations they learn about the game states converge to a similar space. 
This result highlights a level of consistency and robustness in the way generative models process game-related information. Despite differences in architecture or training objectives, the models seem to internalize and represent Othello game states in a similar manner. This convergence suggests that these models, when trained on the Othello task, are not only learning task-specific patterns but are also aligning on a shared understanding of the underlying problem space. To sum up, such alignment enhances the interpretability of these models, as their internal representations become more comparable.

\subsection{Mapping Across Different Layers}
\label{section:mapping}
We compare the mapping similarity across different Decoder hidden layers in GPT-2 and Flan-T5\footnote{We use GPT-2-small and Flan-T5-Base trained on 20k SYNTHETIC dataset, as both have 12 decoder hidden layers.} to understand how each model progressively learns to represent the Othello game, evolving from simple board states to more complex strategies. As shown in Figure \ref{fig:heatmap}, despite their structural differences, GPT-2 and Flan-T5 exhibit similar learned representations at corresponding layers. Both models, when trained on Othello game sequences, seem to converge toward learning comparable internal representations, as highlighted by the heatmap. This conclusion is supported by the following observations: \begin{wrapfigure}{r}{0.5\linewidth}
    \centering
    \vspace{-10pt}
    \includegraphics[width=0.9\linewidth]{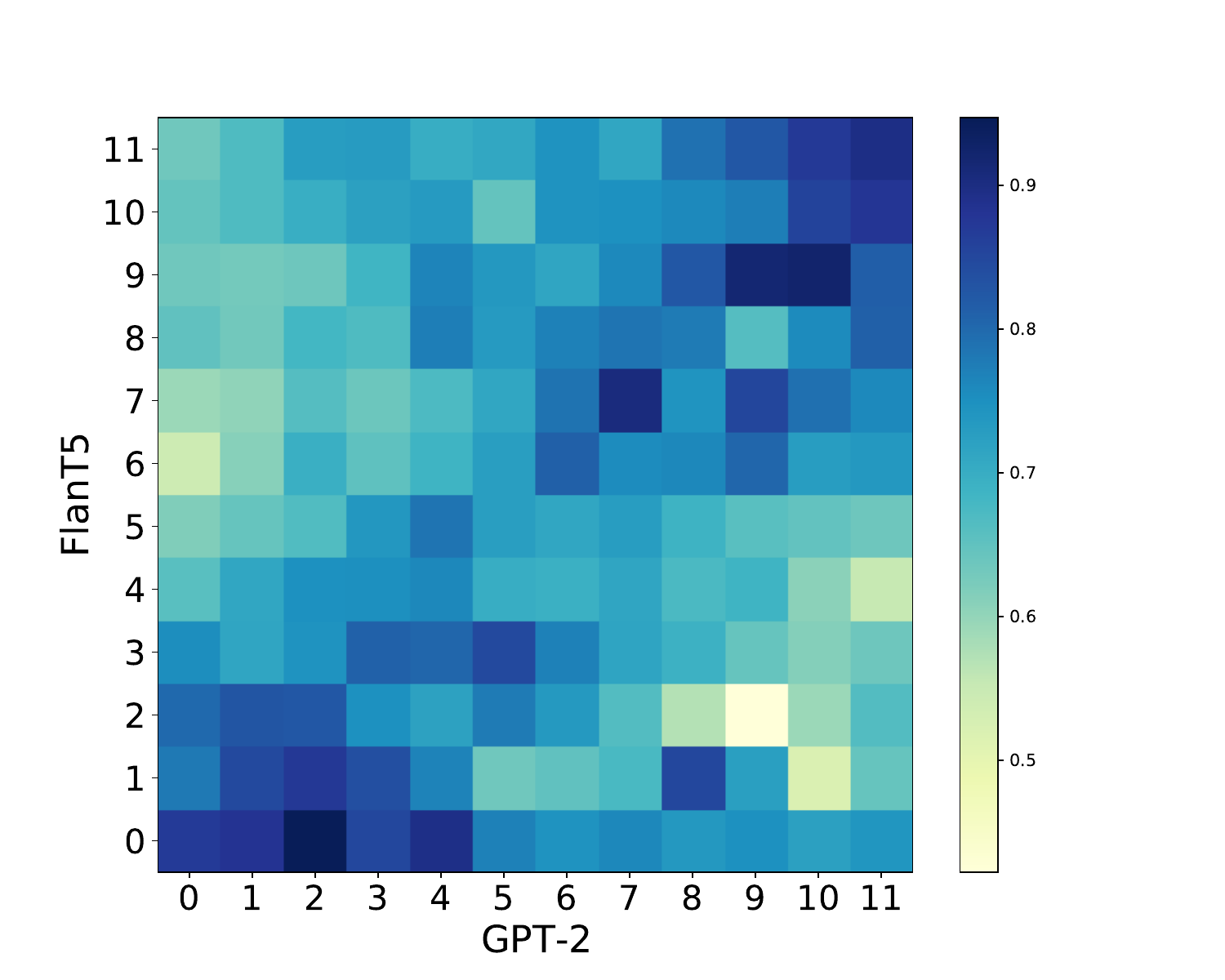}
    \caption{Decoder feature similarity heatmap across different layers.}
    \label{fig:heatmap}
\end{wrapfigure}
(1) \textbf{High Similarity in the Upper Right Diagonal}.
The heatmap reveals a prominent diagonal pattern where corresponding layers from GPT-2 and Flan-T5 show high similarity, especially in the upper half of the heatmap.
This suggests that, despite their differing architectures (GPT-2 being autoregressive and Flan-T5 following an Encoder-Decoder structure), models eventually learn something in common (particularly at layer 11, where high similarity is observed) despite the difference from the beginning. 
This alignment indicates that their layer-wise learning processes evolve in comparable ways as they both adapt to the Othello game environment.
(2) \textbf{Layer-Specific Correspondences}. We notice that specific layers in GPT-2 show high similarity with certain layers in Flan-T5, even though they may not follow a strict diagonal pattern, this suggests that both models are learning certain shared features or patterns in game sequences at particular stages of their processing pipelines.
\begin{figure}[t]
\centering
    \subfigure[\textbf{T5--1 (white)}]
    {
        \includegraphics[width=0.25\textwidth]{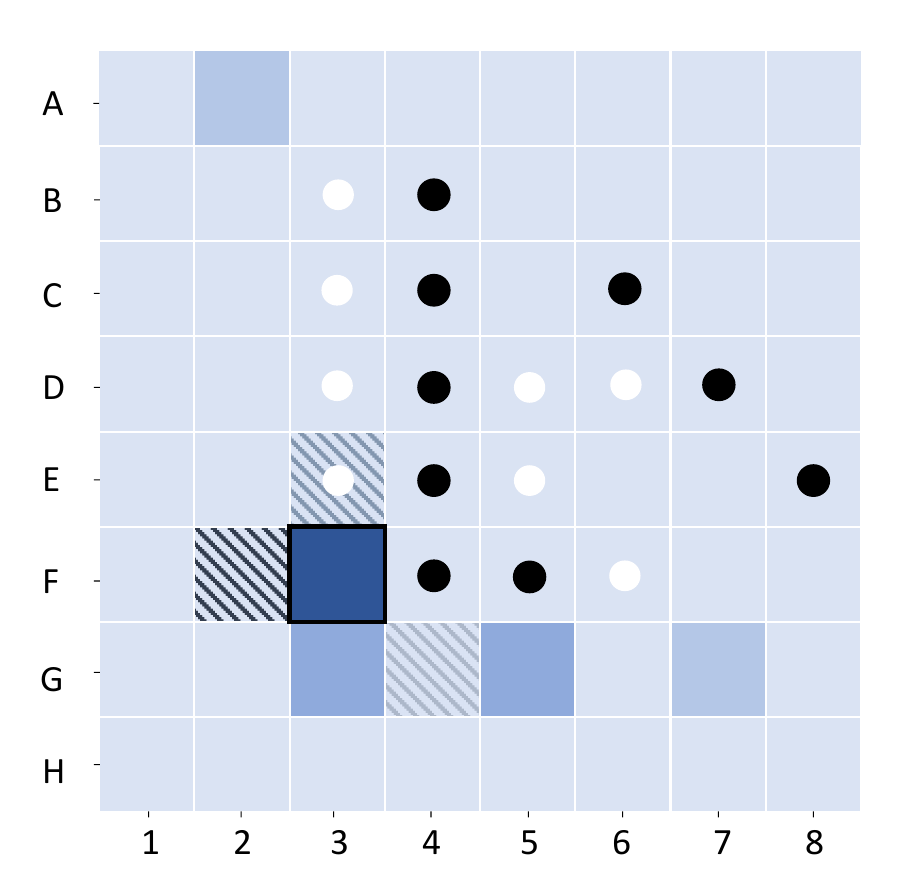}
        \label{fig:first_sub}
    }
    \subfigure[\textbf{T5--2 (black)}]
    {
        \includegraphics[width=0.25\textwidth]{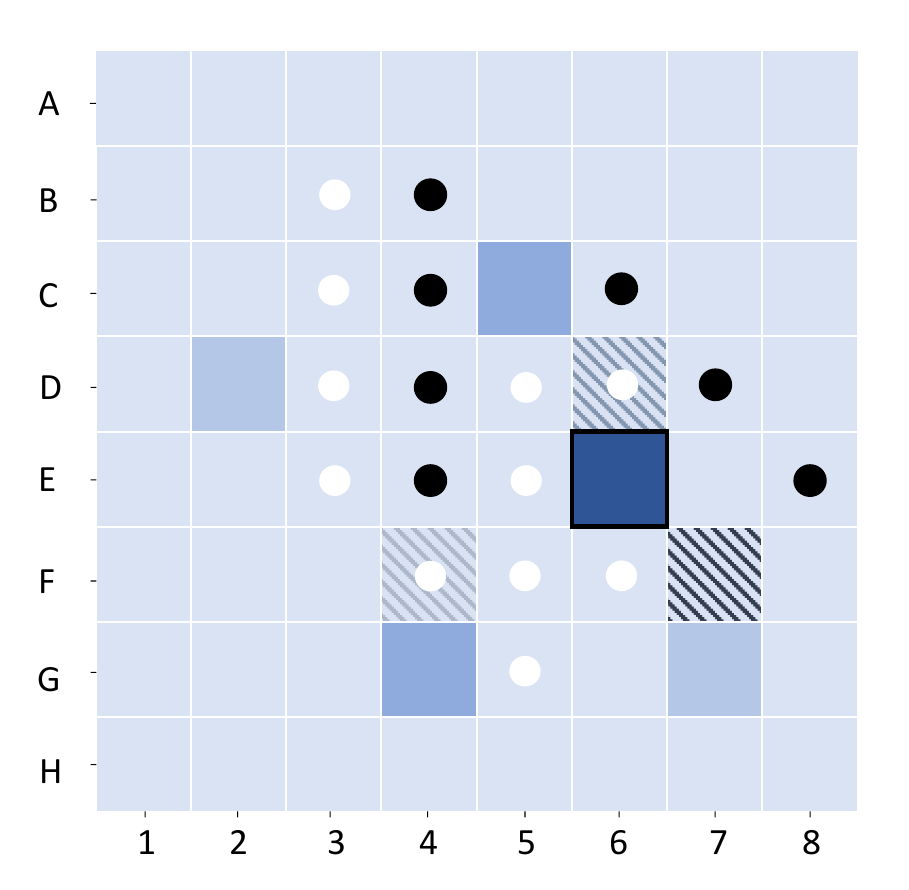}
        \label{fig:second_sub}
    }
    \subfigure[\textbf{T5--3 (white)}]
    {
        \includegraphics[width=0.25\textwidth]{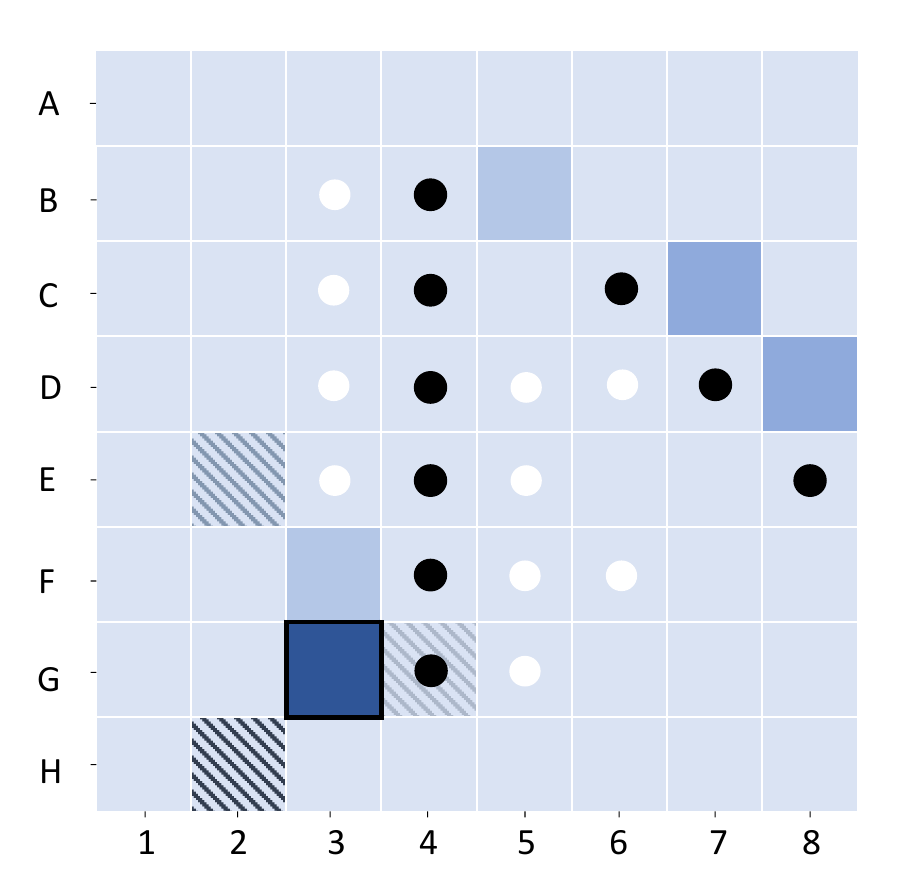}
        \label{fig:third_sub}
    }\\
    \subfigure[\textbf{Mistral--1 (white)}]
    {
        \includegraphics[width=0.25\textwidth]{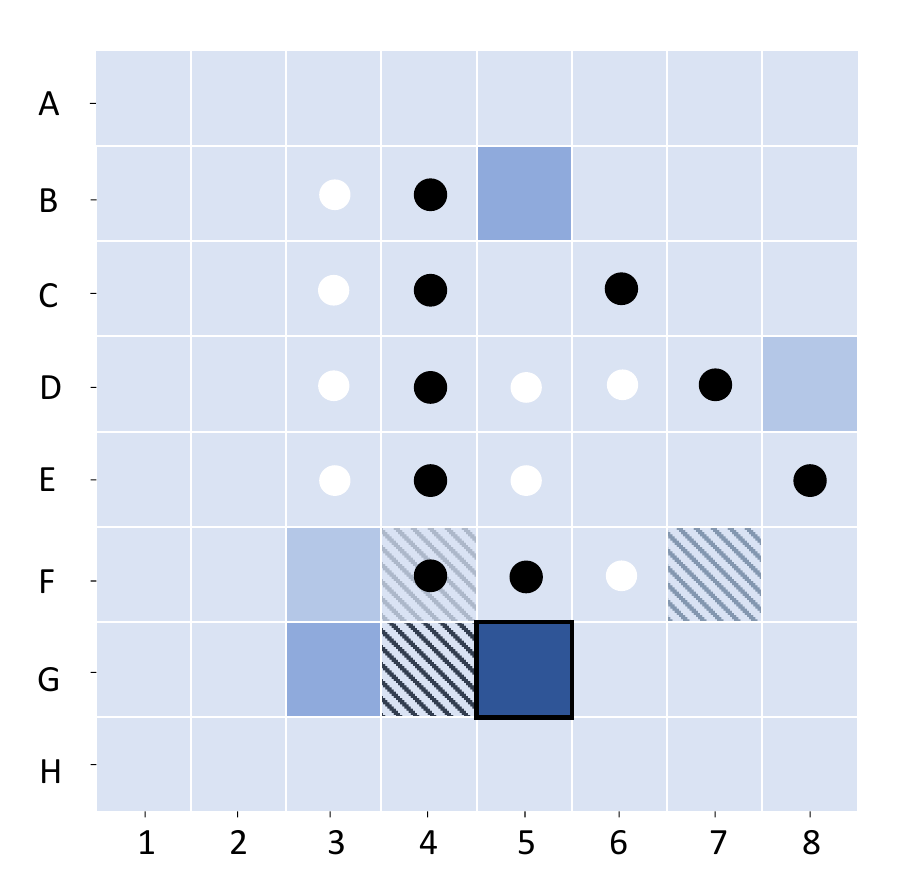}
        \label{fig:fourth_sub}
    }
    \subfigure[\textbf{Mistral--2 (black)}]
    {
        \includegraphics[width=0.25\textwidth]{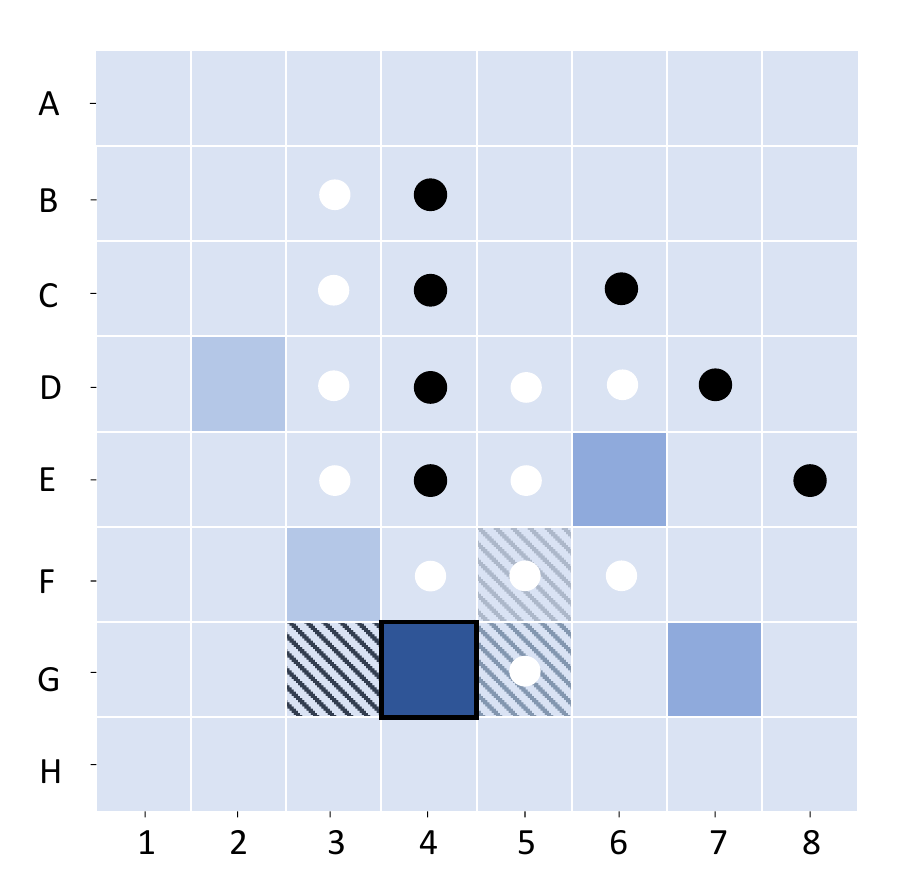}
        \label{fig:fifth_sub}
    }
    \subfigure[\textbf{Mistral--3 (white)}]
    {
        \includegraphics[width=0.25\textwidth]{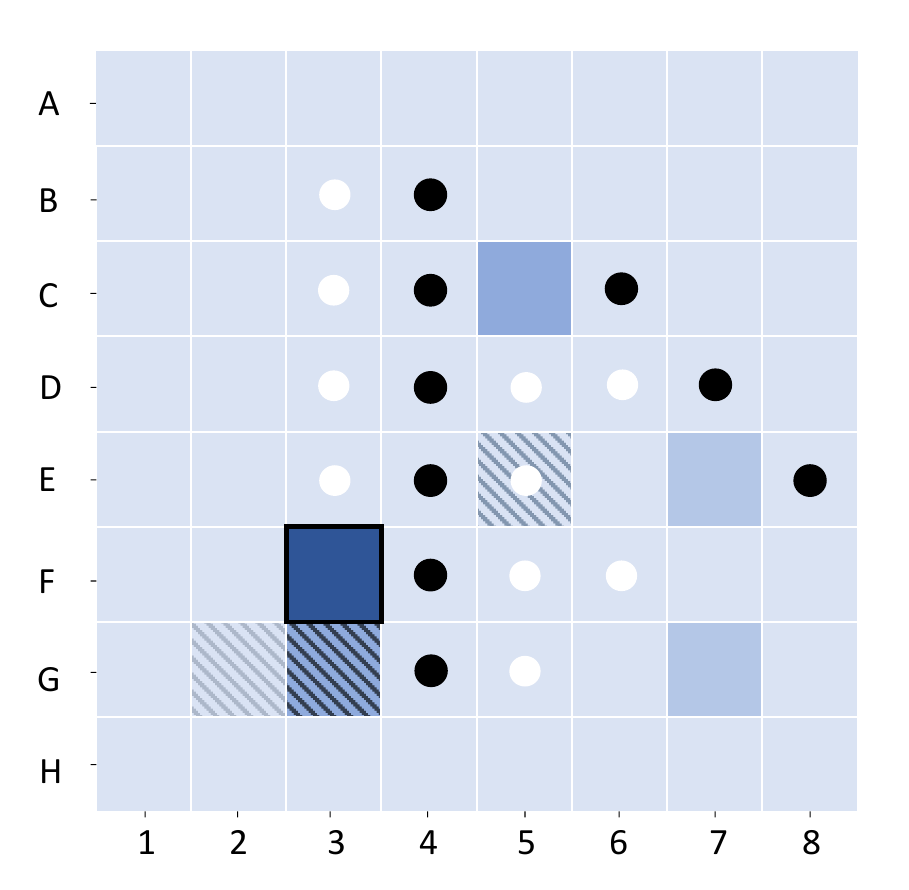}
        \label{fig:sixth_sub}
    }
    \\
    \caption{Othello latent move projection from two best performed models. Colors indicate the likelihood of the position of the next step. Shadows highlight the top three tiles with embeddings closest to the top candidate, with the darkest color in the black box.}
    \label{visanalysis}
\end{figure}
\section{Latent Move Projection: What else does LLMs learn?}
To gain deeper insights into how models learn strategies and predict future moves, we project latent features onto a visual space. For a given game sequence $X_{<i}$, we highlight the top-5 candidate tile positions with the highest predicted probabilities for the next move. Additionally, we compare the embeddings of the top candidate tile with those of the other tiles. We mark the top three tiles whose embeddings are closest to the top candidate to examine their spatial relationships on the board.

We perform latent move projection on the Othello game steps of two models in Figure \ref{visanalysis} (other models see Appendix \ref{latent-full}). Both models successfully predict legal moves given a game sequence, assigning high scores to other legal moves as well (lighter blue tiles). This demonstrates that, with sufficient game sequence data, the model learns the game rules. To further investigate whether the models capture the physical position of each tile, we use shadow marks to highlight the tiles with the closest embedding distance to the tile in the black box. The intensity of the shadow reflects the degree of similarity. We observe that the top-1 tile with the highest similarity (F2 in Figure \ref{fig:first_sub}, G4 in Figure \ref{fig:fourth_sub}) is the one adjacent to the black box tile in both models. This indicates the models not only understand the game mechanics but also capture the spatial relationships between tiles. 
\section{Conclusion}
We conduct a detailed probing of language models' ability to predict legal moves in the Othello board game, based on the settings in~\cite{li2023emergent}. We evaluate seven language models, training them to predict the next move based on previous  moves. All seven models achieve almost `perfect' one-hop move prediction performance when trained with large amount of data. We then adopt representation alignment tools to align the learned game state features from different models into one joint space. We observe high similarity in the board features they learned. In addition, latent move projection is performed to show the models not only understand the game mechanics but also capture the spatial relationships
between tiles. These results, in our view, provide more solid evidence to date of the Othello World Model Hypothesis presented in previous works. Potential impact and future directions are discussed in Appendix \ref{appendix:impact} and \ref{appendix:future}.



\bibliography{iclr2025_conference}
\bibliographystyle{iclr2025_conference}

\appendix
\section{Othello World Model Hypothesis}
According to previous works~\cite{li2023emergent,Nanda2023EmergentLR}, a world model refers to a representation or a mapping of a world, ideally a homomorphism. Language models have been shown to develop internal representations for simple concepts, such as color and direction~\citet{Patel2022MappingLM,Abdou2021CanLM}. Training language models on Othello game sequences further supports the idea that LLMs can function as a world model. This is demonstrated by their ability to learn and internalize the structured dynamics and rules of a complex system, rather than merely memorizing patterns. This capacity highlights their potential for understanding and representing intricate environments through abstract, systematic reasoning.
\label{worldmodel}
\section{Dataset Statistics}
\label{datasetstatistics}
The details of the two datasets are listed in Table \ref{tab:datastatistics}.
 \input{tables/datastatistics} 
\section{Compared Methods}
\label{sec:appendix2}
We perform our experiments using several existing baselines, with both Encoder-Decoder or Decoder-only structures. We first adopt some popular language models such as

\textbf{GPT-2.} We fine-tune GPT-2 to generate the whole game sequence step by step. Specifically, we use the smallest version of GPT-2.

\textbf{Bart.} We use Bart-base to generate the sequence by feeding the first token into the Encoder and fine-tuning the model to generate the remaining tokens.

\textbf{T5.} Similar as Bart, we adopt \verb|T5-base| in our experiment.

We then adopt several LLMs for the task:

\textbf{Flan-T5.} We adopt \verb|Flan-T5-XL|, which contains 3B parameters in our experiment.

\textbf{LLaMA-2.} We use \verb|LlaMa2-7B| and only fine-tune the LoRA adapter in our experiment.

\textbf{Mistral.} We use \verb|Mistral-7B| in our experiments. Similar to LLaMA-2, we also only fine-tune the LoRA adapter but keep the rest of parameters fixed.

\textbf{Qwen2.5.} We use \verb|Qwen2.5-7B| in our experiments, one of the most state-of-the-art LLMs for sequence generation.
\section{Model Size Analysis on Two-hop Generation}
\begin{figure}[ht!]
\centering
    \subfigure[\textbf{CHAMPIONSHIP}]
    {
        \includegraphics[width=0.4\textwidth]{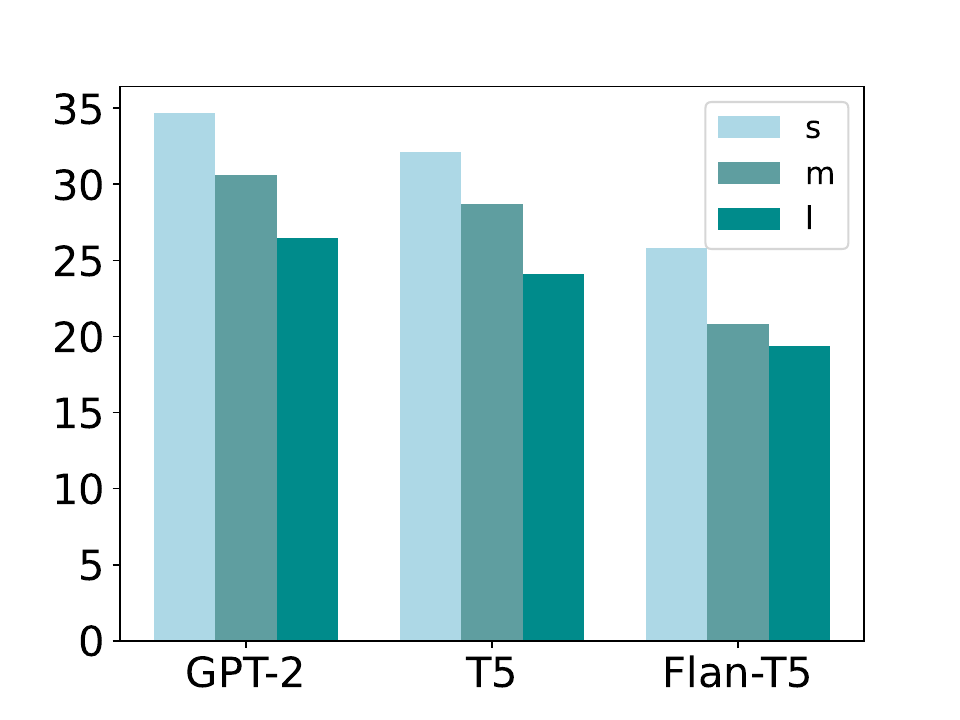}
        \label{fig:second_sub}
    }
    \subfigure[\textbf{SYNTHETIC}]
    {
        \includegraphics[width=0.4\textwidth]{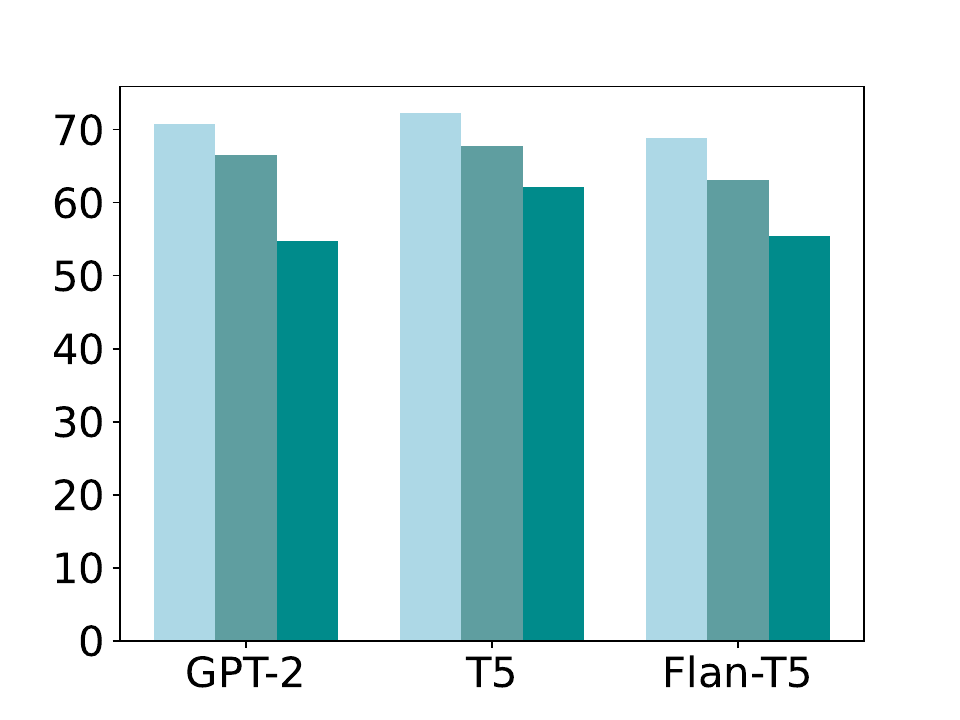}
        \label{fig:third_sub}
    }
    \caption{Othello 2-hop generation performance under different model sizes. All models are non-pretrained versions fine-tuned with 20k game sequences.}
    \label{modelsizeanalysis1}
\end{figure}

We present the 2-hop performance across various model sizes in Figure \ref{modelsizeanalysis1}. As we scale up the model, the error rate decreases, suggesting that a larger model size positively affects game understanding. However, the impact of model size diminishes when compared to the 1-hop performance, indicating a diminishing return on performance gains with increased model size.

\section{Data Size Analysis on CHAMPIONSHIP Dataset}
\begin{figure}[!ht]
    \centering
    \subfigure[\textbf{Non-pretrained}]
    {
        \includegraphics[width=0.43\textwidth,height=42mm]{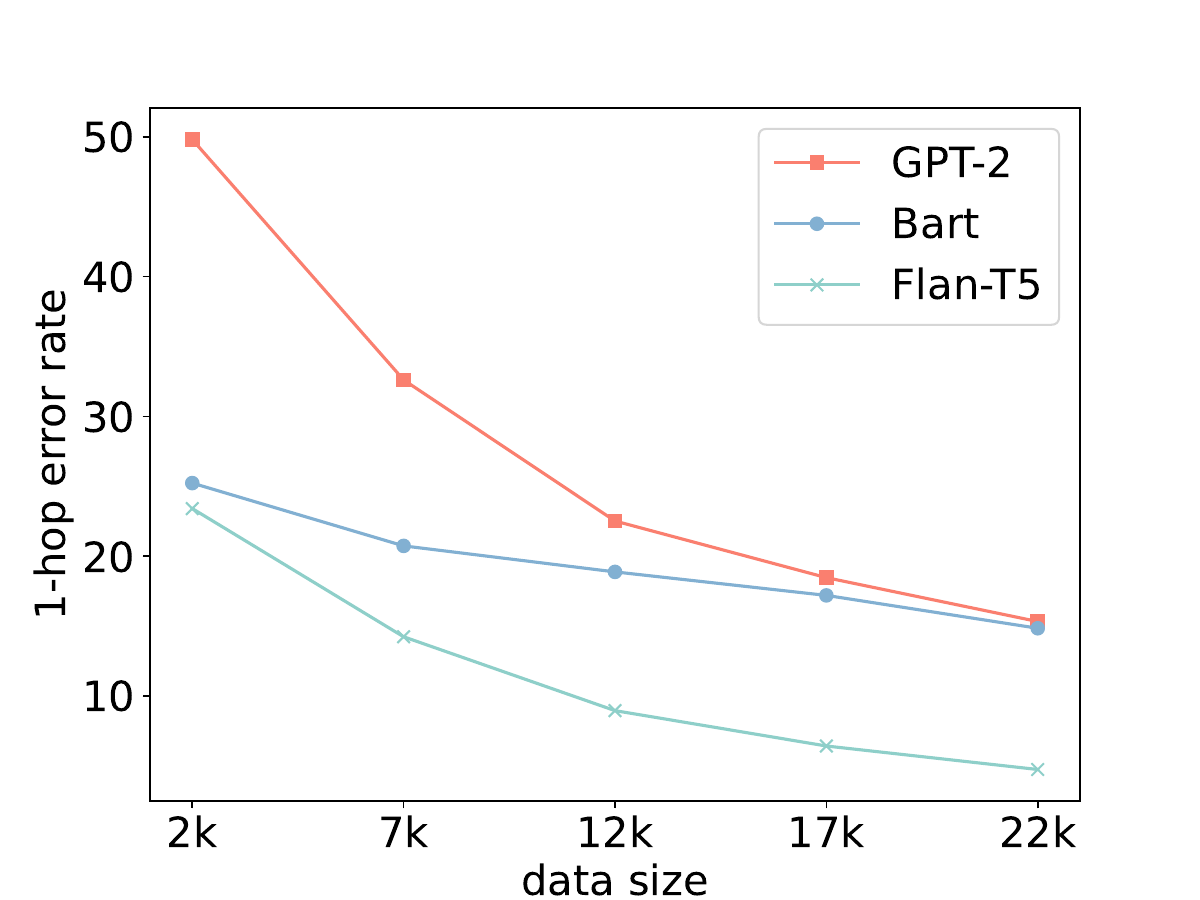}
        \label{fig:second_sub}
    }
    \subfigure[\textbf{Pretrained}]
    {
        \includegraphics[width=0.43\textwidth,height=42mm]{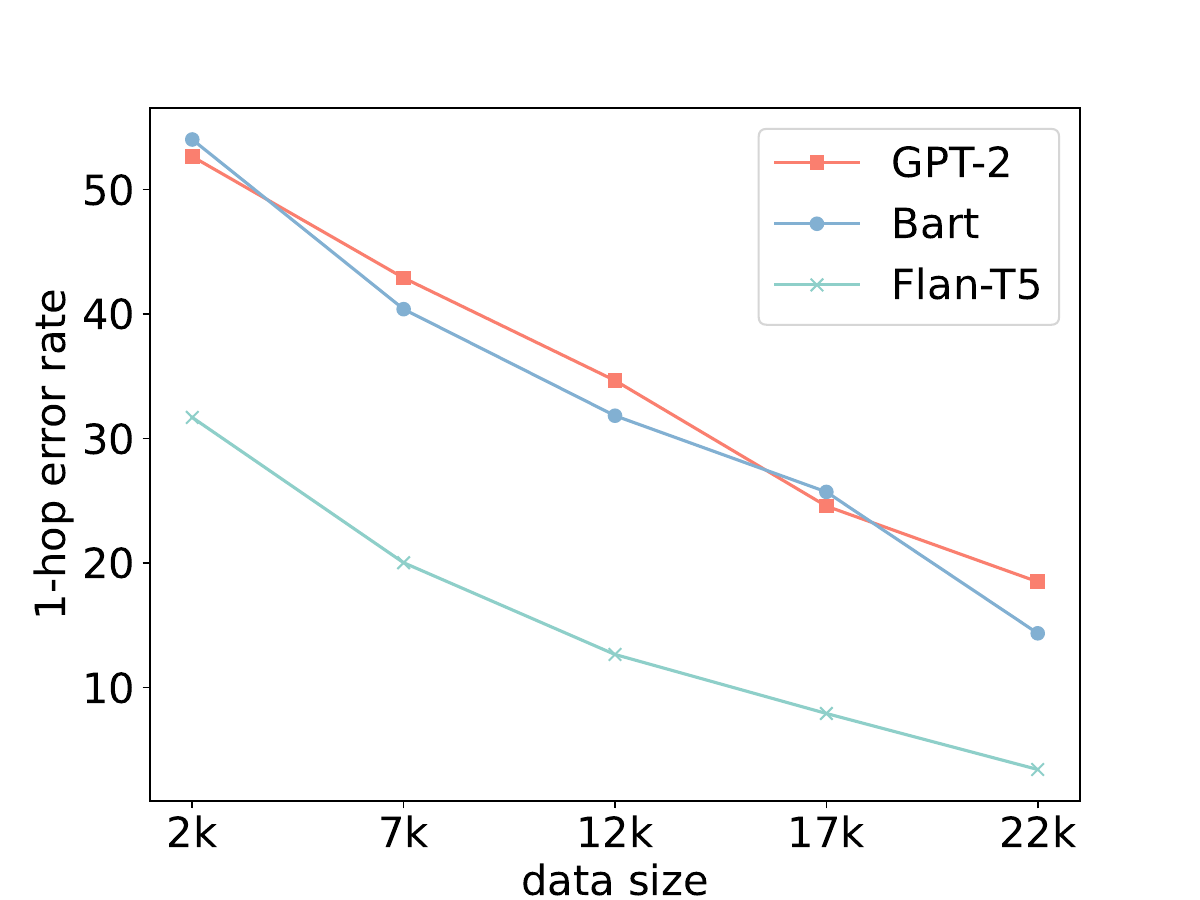}
        \label{fig:third_sub}
    }
    \caption{Analysis of 1-hop error rates on the CHAMPIONSHIP dataset with varying data scales.}
    \label{sizeanalysis1}
\end{figure}

We also present the data size analysis on the CHAMPIONSHIP dataset (see Figure \ref{sizeanalysis1}). We see similar conclusions as in Figure \ref{sizeanalysis}. The prediction accuracy gets better when we increase the data size. Also, the error rate demonstrates a more steady drop in models pretrained with upstream language modeling tasks.
\section{Supervised Mapping Heatmap}
\begin{figure}[ht!]
    \centering \includegraphics[width=0.5\linewidth]{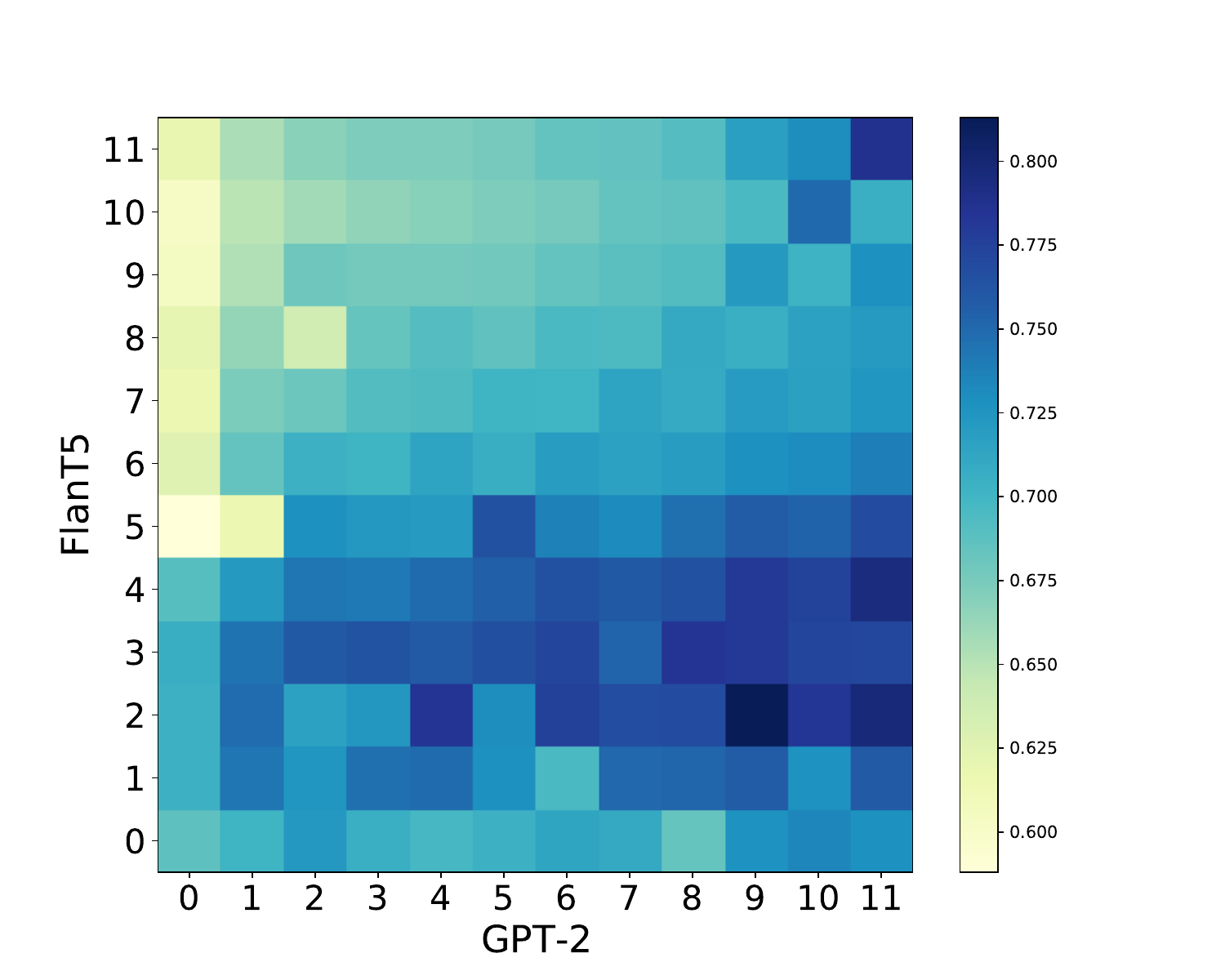}
    \caption{Decoder feature similarity (supervised) heatmap across different layers.}
    \label{fig:heatmap1}
\end{figure}

We also present the supervised mapping results for the same setting in Section \ref{section:mapping}. The mapping in Figure \ref{fig:heatmap1} reveals a more pronounced diagonal pattern of similarity, with particularly high similarity observed in the upper-right corner. This provides further evidence that the models converge and acquire shared knowledge when trained on Othello data, indicating a strong alignment in their learned representations.
\section{Latent Move Projection (Full Version)}
\label{latent-full}
\begin{figure}[t]
\centering
    \subfigure[\textbf{Bart--1 (white)}]
    {
        \includegraphics[width=0.3\textwidth]{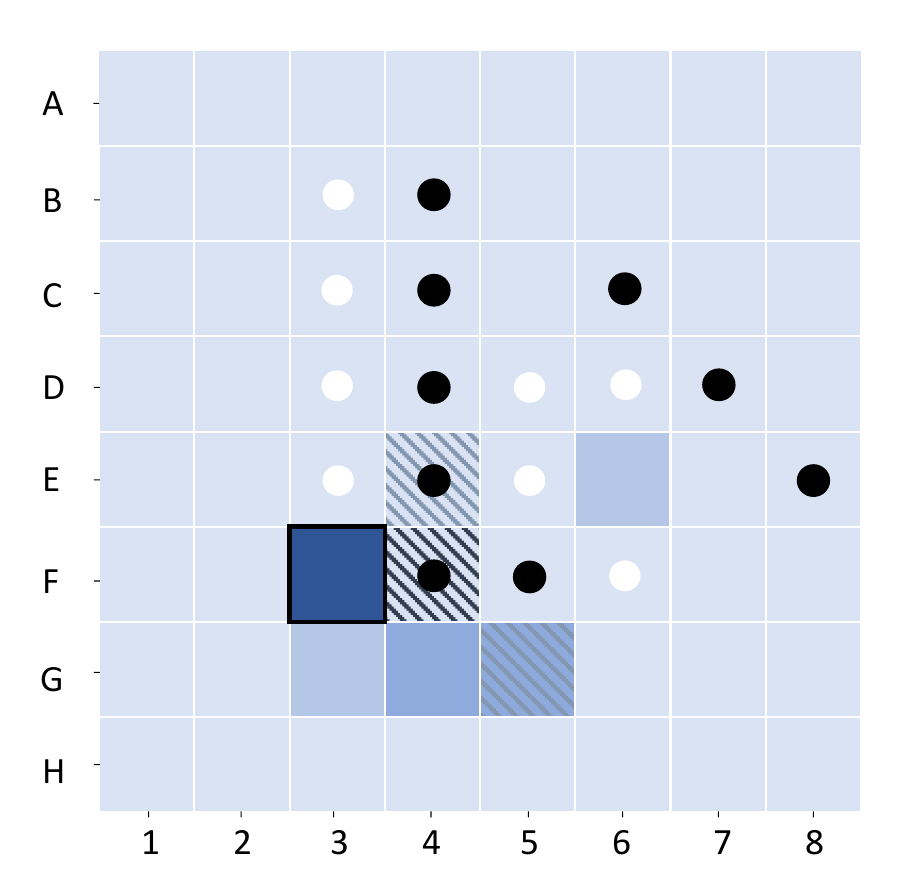}
        \label{fig:first_sub}
    }
    \subfigure[\textbf{Bart--2 (black)}]
    {
        \includegraphics[width=0.3\textwidth]{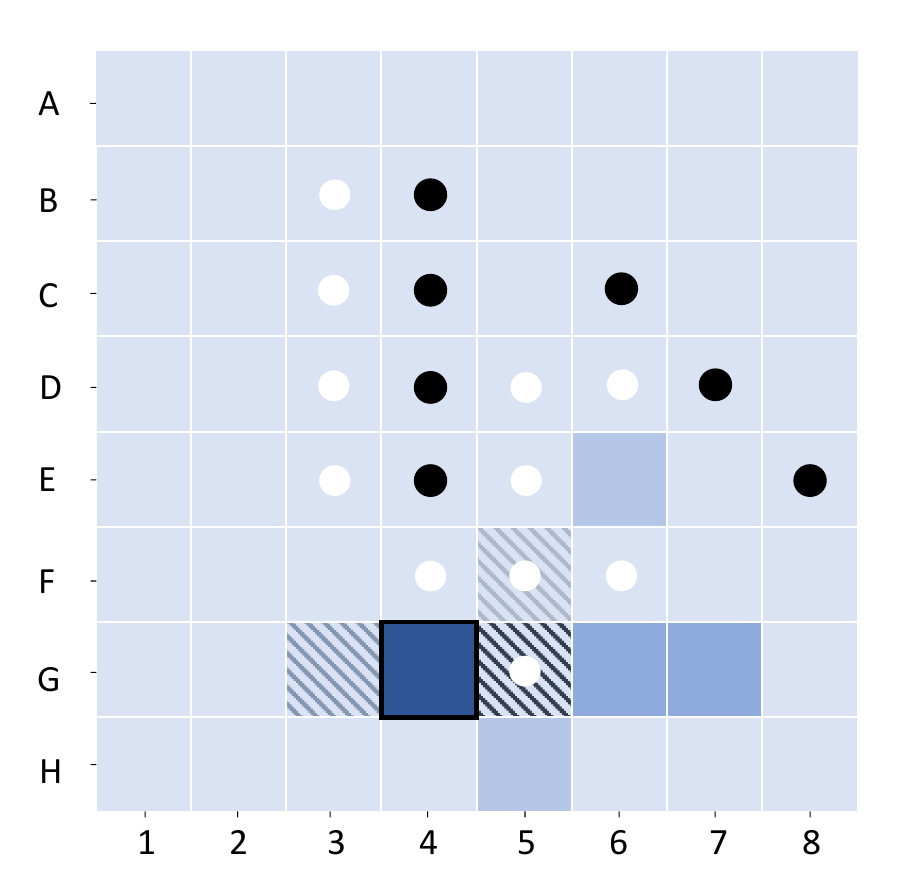}
        \label{fig:second_sub}
    }
    \subfigure[\textbf{Bart--3 (white)}]
    {
        \includegraphics[width=0.3\textwidth]{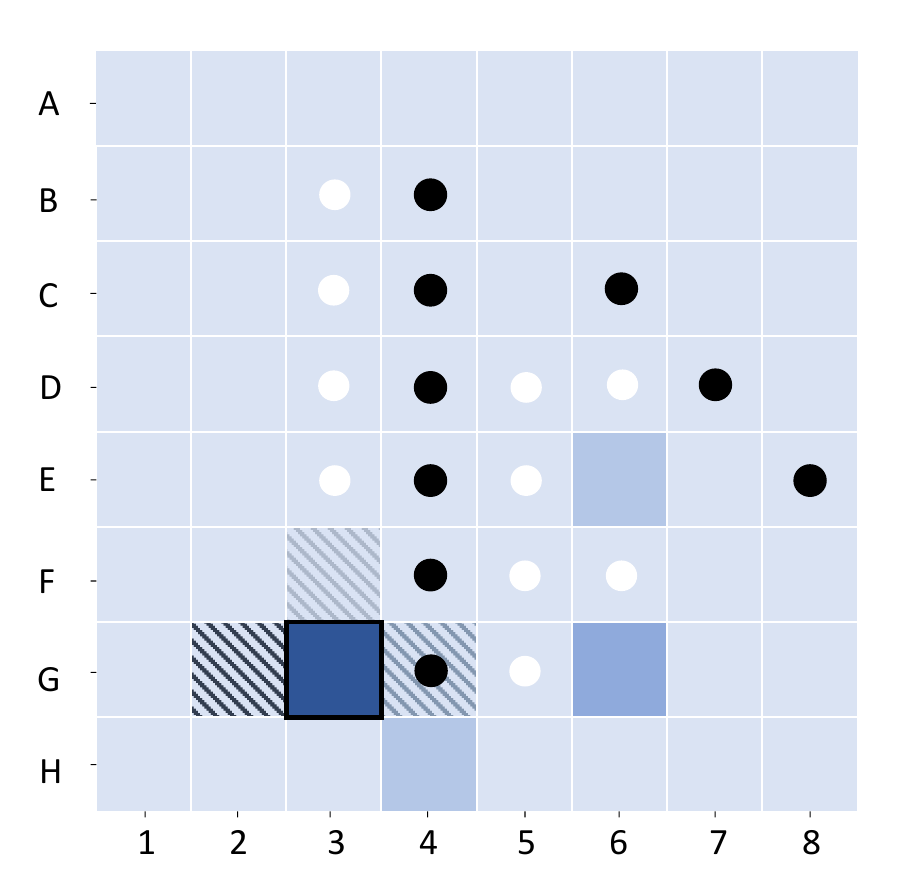}
        \label{fig:third_sub}
    }\\
    \subfigure[\textbf{GPT2--1 (white)}]
    {
        \includegraphics[width=0.3\textwidth]{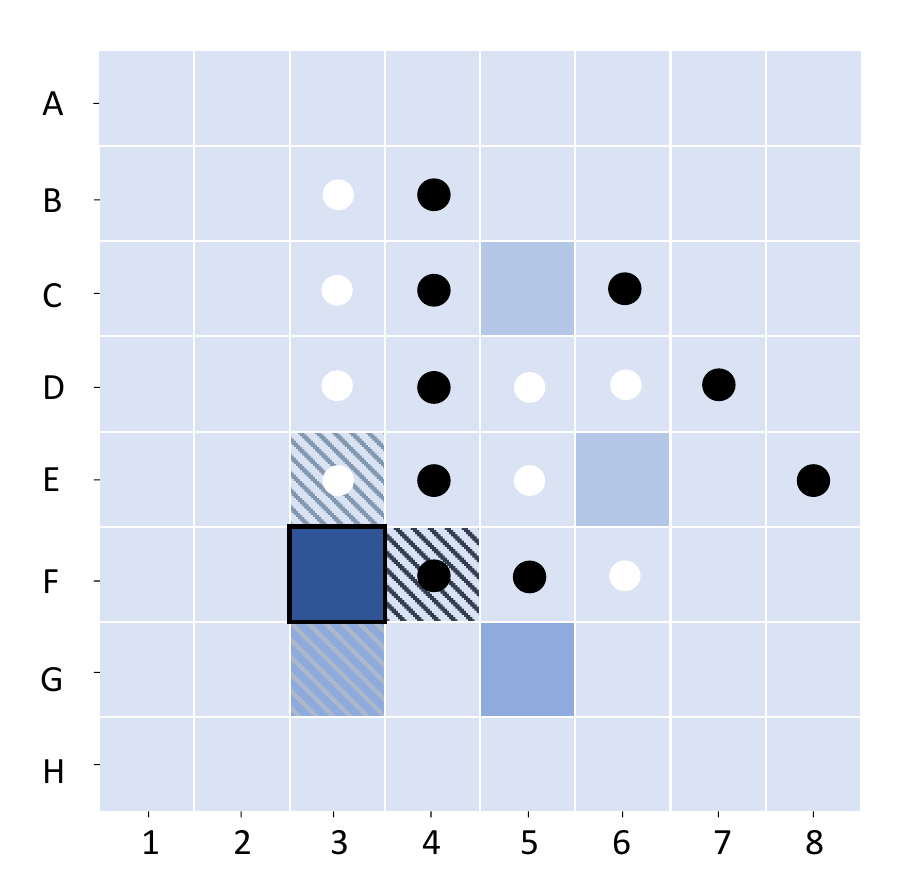}
        \label{fig:fourth_sub}
    }
    \subfigure[\textbf{GPT2--2 (black)}]
    {
        \includegraphics[width=0.3\textwidth]{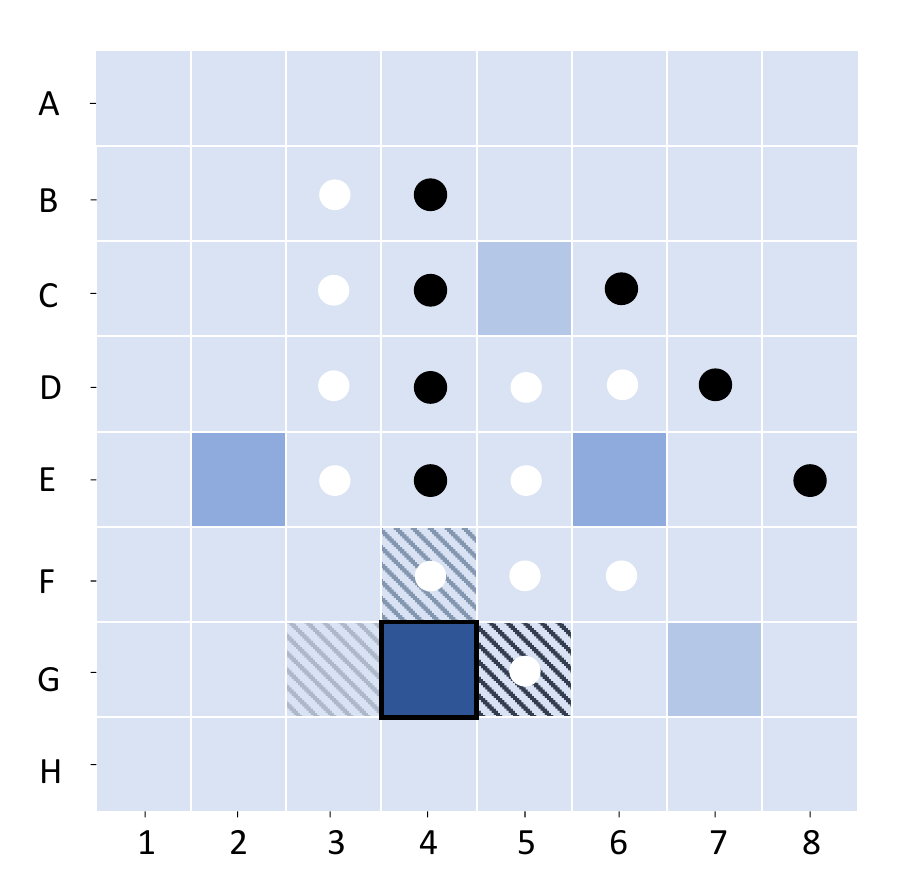}
        \label{fig:fifth_sub}
    }
    \subfigure[\textbf{GPT2--3 (white)}]
    {
        \includegraphics[width=0.3\textwidth]{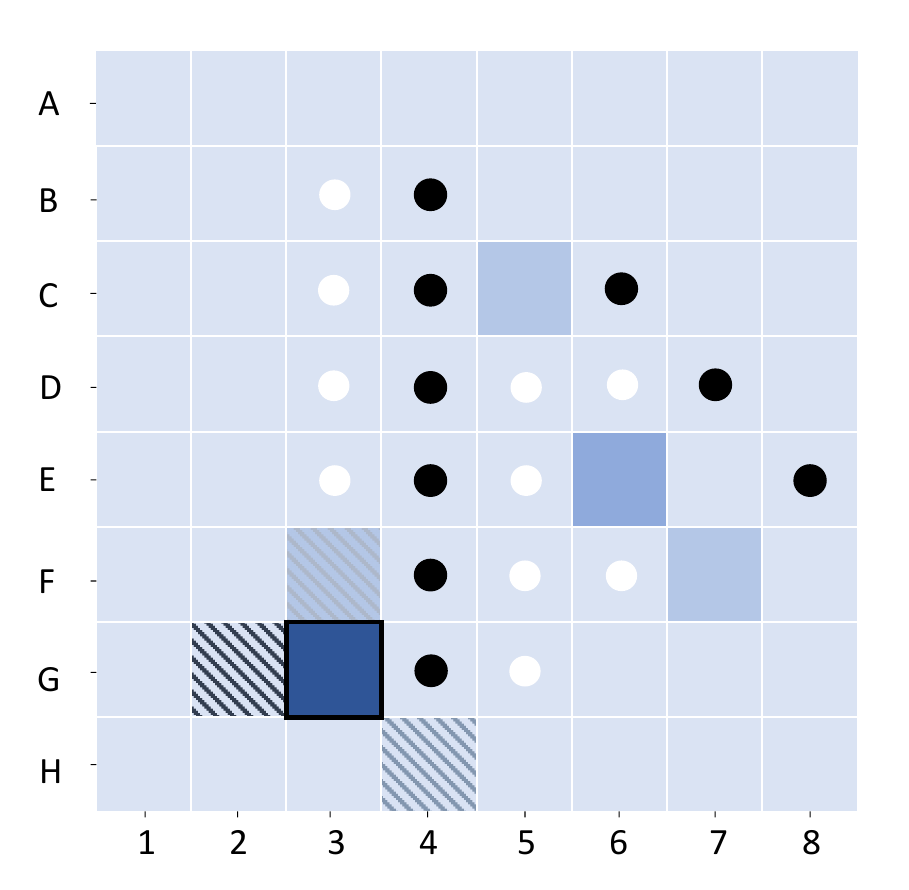}
        \label{fig:sixth_sub}
    }
    \\
    \subfigure[\textbf{LlaMa2--1 (white)}]
    {
        \includegraphics[width=0.3\textwidth]{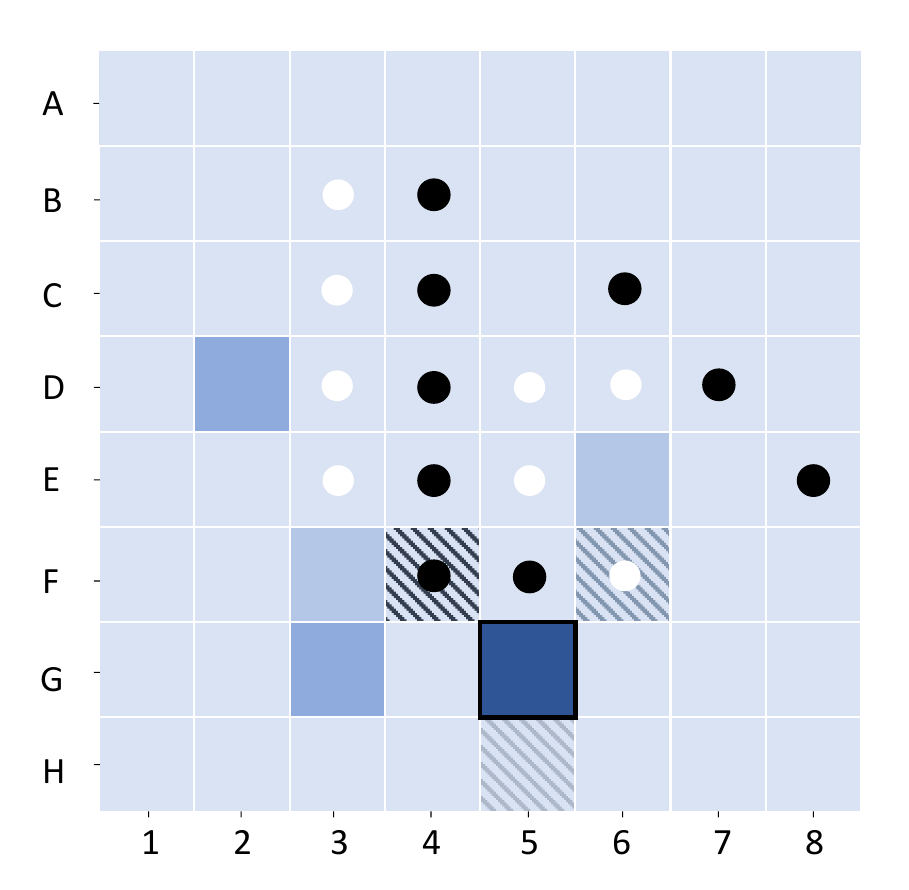}
        \label{fig:fourth_sub}
    }
    \subfigure[\textbf{LlaMa2--2 (black)}]
    {
        \includegraphics[width=0.3\textwidth]{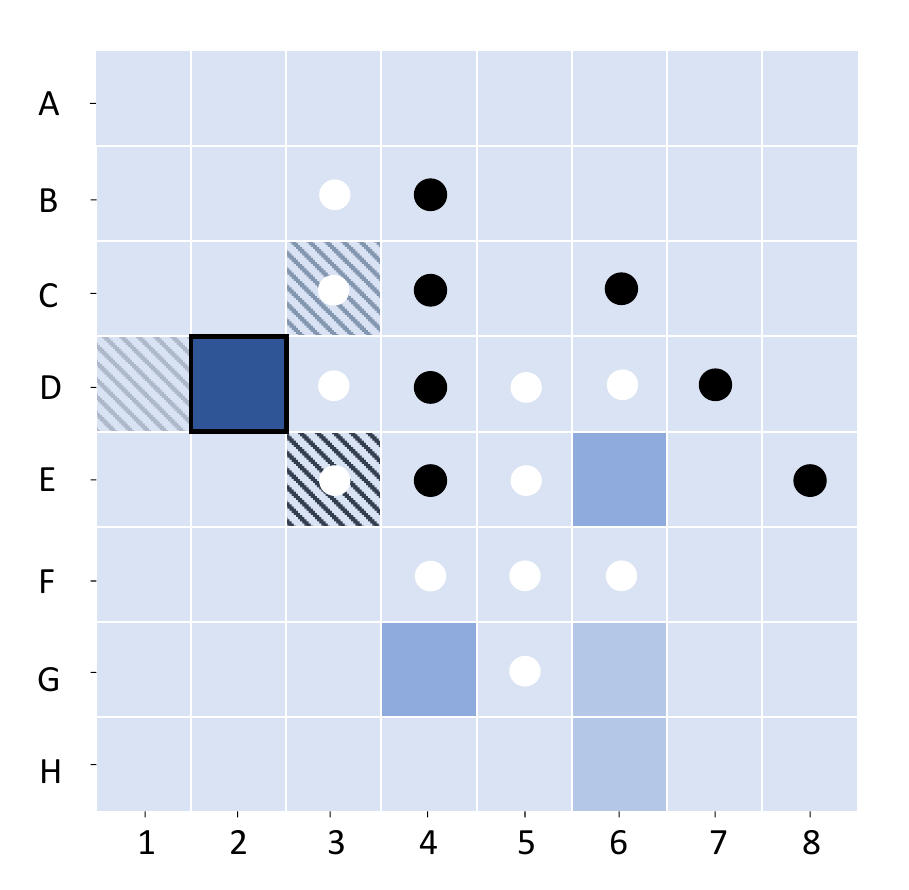}
        \label{fig:fifth_sub}
    }
    \subfigure[\textbf{LlaMa2--3 (white)}]
    {
        \includegraphics[width=0.3\textwidth]{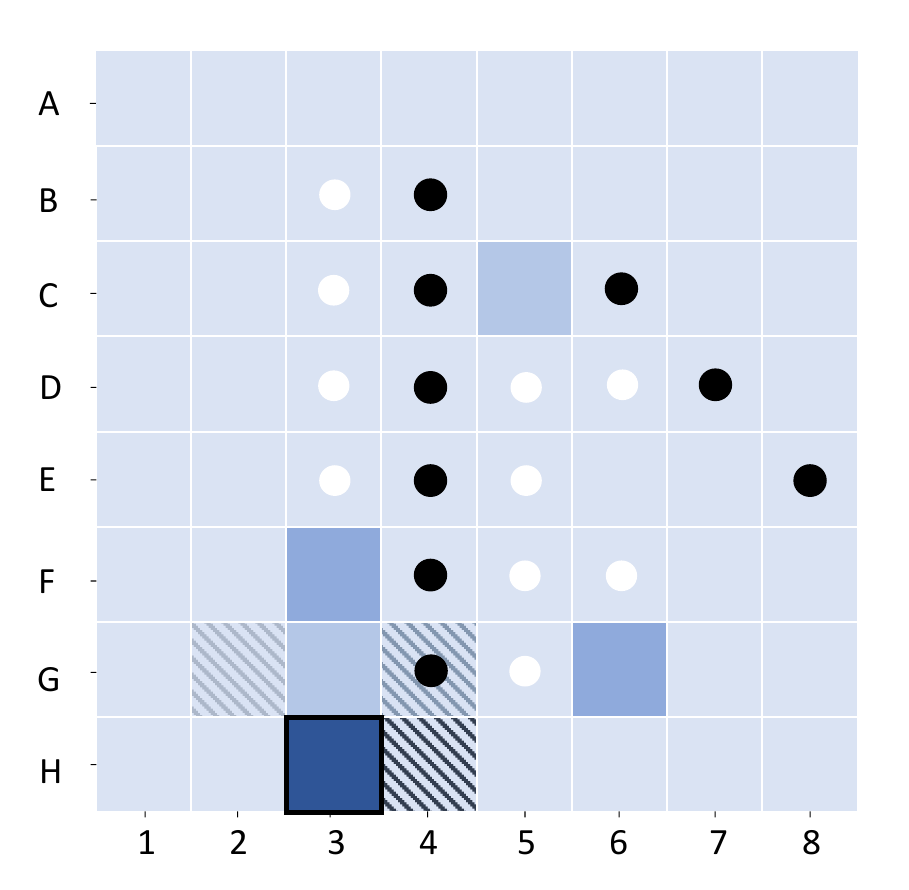}
        \label{fig:sixth_sub}
    }
    \\
    \caption{Othello latent move projection from two best performed models.}
    \label{visanalysis_full}
\end{figure}
We attach the prediction from different models of the same game state in Figure \ref{visanalysis_full}. By comparing the performance of different models on the task, we find that overall, Mistral shows the best performance. It consistently demonstrates the best performance across different scenarios, effectively generating legal moves and showing a nuanced understanding of game rules. The Bart model frequently predicts adjacent tiles, leading to numerous failure cases, particularly when trained with smaller datasets. Llama-2 exhibits inconsistent performance, with a tendency to favor certain tile positions or exhibit a bias in move selection. While its predictions are often reasonable, the model appears to lack the robust policy understanding seen in Mistral, especially under constrained training conditions.
\section{Limitations}
Although this work demonstrates the ability of different language models to understand Othello game rules, several limitations persist that require further investigation:

\textbf{Challenges in Multi-step Move Generation.} 
While language models can predict the next move with reasonable accuracy, they struggle to predict entire game sequences. The key question is whether strong multi-step performance is a reasonable expectation. Othello is a dynamic game where optimal play often involves sacrificing short-term gains for long-term advantages. The complexity arises from the interplay of distinct player strategies and the rotational invariance of the board, leading to many game states where the subjectively or objectively best move is inherently underdetermined. As a result, the ability to accurately predict entire sequences may remain elusive, given the complexity and variability of decision-making in the game.

\textbf{Limitations in Data Requirements.} 
Our experiments show that reducing the 1-hop error rate to less than 0.1\% 
demands a large volume of training data. This reliance on vast datasets presents a scalability issue, as access to Othello game data is limited. Moreover, training on such large datasets is computationally expensive and time-consuming, which can be a prohibitive factor for many researchers or organizations without access to substantial computational resources.


\section{Potential Impact}
\label{appendix:impact}
Training language models on Othello game sequences can imply that LLMs function as a world model because it showcases their ability to learn and internalize the structured dynamics and rules of a complex system, rather than merely memorizing patterns. Investigating the parallels between how language models learn structured representations and how humans internalize similar concepts can shed light on the cognitive processes underlying reasoning, strategy, and language. This could deepen our understanding of human cognition and inform theories of learning and representation. The observation that language models, regardless of architecture or scale, learn similar patterns from Othello game sequences suggests that these models converge on universal representations when trained on structured data. This implies that the underlying mechanisms of representation learning in LLMs are robust and consistent, highlighting their ability to capture the rules and dynamics of structured systems. The ability of language models to learn patterns from Othello sequences provides more hints on the idea that LLMs can act as world models, capable of internalizing rules, strategies, and dynamics. This has far-reaching implications for tasks requiring reasoning about complex environments, such as planning, simulation, and autonomous decision-making.
\section{Future Directions}
\label{appendix:future}
We list several possible future directions to study how our results could generalize to other broader scenarios.

\textbf{More Complicated Games.} Since this work is primarily limited to the Othello game, an intriguing question arises: could similar findings be observed in other games such as chess, checkers, or Go? These games, like Othello, involve strategic planning, dynamic state transitions, and trade-offs between short-term gains and long-term advantages. Exploring how large language models (LLMs) learn and represent strategies in these contexts could be highly valuable.

\textbf{Multimodal Support.} Leveraging Multimodal LLMs (MLLMs) to train models and investigate feature alignment across different modalities is also a highly relevant and promising research direction. In the context of Othello, this approach could involve aligning visual representations of the game board with text-based sequencial moves. Such alignment can help bridge the gap between symbolic reasoning and natural language understanding, enabling models to not only predict optimal moves but also provide hints if the world model theory could also be applied in other modalities.

\end{document}

%% file: math_commands.tex
\usepackage{amsmath,amsfonts,bm}

\def\eqref#1{equation~\ref{#1}}

\def\1{\bm{1}}

\DeclareMathAlphabet{\mathsfit}{\encodingdefault}{\sfdefault}{m}{sl}
\SetMathAlphabet{\mathsfit}{bold}{\encodingdefault}{\sfdefault}{bx}{n}



%% file: tables/mainexp.tex
\begin{table*}[]
\centering
\small
\begin{adjustbox}{max width=0.99\textwidth}
    {
\begin{tabular}{p{1.3cm}p{0.7cm}cp{0.7cm}p{0.7cm}p{0.7cm}cp{0.7cm}p{0.7cm}p{0.7cm}p{0.7cm}p{0.8cm}}
\hline
\multirow{ 2}{*}{Method} & \multirow{ 2}{*}{Type} & \multirow{ 2}{*}{P} &\multicolumn{3}{c}{\textbf{CHAMPIONSHIP}} & &\multicolumn{5}{c}{\textbf{SYNTHETIC}} \\ \cline{4-6} \cline{8-12} 
     &  &  & 2k       & 20k       & full      &  & 2k  & 20k  & 200k  & 2M & full \\ 
\hline

  GPT-2  & D  & \XSolidBrush  &  49.8        &       17.7  &    5.6 & & 49.2     &   26.8   &  \textbf{13.6}    &  10.4  &  $<$0.1    \\
 Bart     & E-D         &    \XSolidBrush       &    25.2     &  16.6&  4.7   &       &  73.6    &  31.7  & 14.2  & 16.3  & $<$0.1 \\
 T5     &    E-D      &     \XSolidBrush      &  \textbf{20.9}           & 15.2   &    4.3   &       & 65.8   & 28.7  & 15.7  & 10.1 &  $<$0.1    \\
 Flan-T5      &  E-D        &   \XSolidBrush        &     23.4      &  \textbf{4.8}    &   \textbf{3.6}    &       &  \textbf{35.6} & \textbf{23.7}   & 21.2  & \textbf{7.7} & $<$0.1 \\
  LlaMa-2      &    D      &   \XSolidBrush     &   27.8         &  16.5    &   5.7  &       &  57.1   &  35.4 & 16.9 & 10.2 & $<$0.1 \\
Mistral     &    D      &    \XSolidBrush       &   22.1     &  14.8    &    4.2   &       &   48.2  & 34.4  & 17.7  & 8.3  & $<$0.1 \\
Qwen2.5     &    D      &    \XSolidBrush       &   25.2     &  17.3    &    5.5   &       &   45.9  & 37.8  & 20.1  & 9.2  & $<$0.1 \\
\hline
   GPT-2  & D  & \Checkmark  &  52.6      &      19.7   &    13.6      & & 74.4    &   32.4   &  19.9    & 14.1  &  $<$0.1  \\
 Bart     & E-D         &    \Checkmark       &   54.0       &  14.6    &   13.7    &       &  77.2   &  35.8 & 24.4  & 16.6  & $<$0.1   \\
 T5    &    E-D      &  \Checkmark         &   45.5        &  19.6     &    3.8   &       & 69.4    & 36.9  & 32.6  & 13.9 & $<$0.1   \\
 Flan-T5      &  E-D        &  \Checkmark         &       31.7   & \textbf{4.8}  &  3.7    &       &   70.3  &   \textbf{25.4}&  45.0  & 8.7  & $<$0.1 \\
  LlaMa-2      &    D      &  \Checkmark         &  43.1     & 14.7    &  7.0    &       &  74.6  &  41.5   & 33.4  & \textbf{7.6}  & $<$0.1  \\
Mistral     &    D      &     \Checkmark      &  \textbf{16.8}          & 15.0    &   \textbf{3.3}   &       &  \textbf{33.8}   &   30.6  & \textbf{18.2}  & 7.7 & $<$0.1  \\
Qwen2.5     &    D      &     \Checkmark      &  20.9          & 18.2    &   6.0   &       &  46.5   &   39.3  & 23.4  & 10.8 & $<$0.1  \\
\hline
\end{tabular}}
\end{adjustbox}
\caption{The error rate (\%) of 1-hop game move generation in terms of different size of training data. `Type' refers to the model type, `P' denotes if the model is pretrained with upstream language modeling tasks or not. Numbers in bold represent best-performing models. }
\label{exp0}
\end{table*}

%% file: tables/mainexp1.tex
\begin{table*}[]
\small
\centering
\begin{adjustbox}{max width=0.99\textwidth}
    {
\begin{tabular}{p{1.3cm}p{0.7cm}cp{0.7cm}p{0.7cm}p{0.7cm}cp{0.7cm}p{0.7cm}p{0.7cm}p{0.7cm}p{0.8cm}}
\hline
\multirow{ 2}{*}{Method} & \multirow{ 2}{*}{Type} & \multirow{ 2}{*}{P} &\multicolumn{3}{c}{\textbf{CHAMPIONSHIP}} & &\multicolumn{5}{c}{\textbf{SYNTHETIC}} \\ \cline{4-6} \cline{8-12} 
     &  &  & 2k       & 20k       & full      &  & 2k  & 20k  & 200k  & 2M & full \\ 
\hline

  GPT-2  & D  & \XSolidBrush  &  78.5       &      34.7   &    28.1      & & 76.3    &   70.8  &  \textbf{43.6}   &  29.0 &  5.2   \\
 Bart     & E-D         &    \XSolidBrush       &   54.2       &  31.1   &  23.4    &       & 86.5   &  67.2 & 44.8 & 35.7 & 4.2 \\
 T5     &    E-D      &     \XSolidBrush      &  \textbf{48.8}          & 28.7  &    24.4  &       & 88.2   & 67.7 & 46.9 & 35.9 &  3.4     \\
 Flan-T5      &  E-D        &   \XSolidBrush        &     51.8      &  \textbf{20.8}   &  \textbf{21.9}   &       &  79.6  & \textbf{63.1}  & 48.6  & 26.7 & \textbf{2.8}\\
  LlaMa-2      &    D      &   \XSolidBrush     &  60.9        &  36.3    &   26.4   &       &  87.3  & 67.8 & 45.2  & 36.3 & 5.5\\
Mistral     &    D      &    \XSolidBrush       &   51.4      &  31.7   &    22.3  &       &  \textbf{71.2}  &  77.1   & 47.9 & \textbf{26.4} & 3.0\\
Qwen2.5     &    D      &    \XSolidBrush       &   55.9      &  25.4   &    22.8  &       &  77.6  &  65.3   & 44.2 & 28.7 & 3.3\\
\hline
   GPT-2  & D  & \Checkmark  &  92.2       &      43.4  &    37.2      & & 99.6    &   \textbf{72.6}  &  45.5   & 34.4 &   6.2  \\
 Bart     & E-D         &    \Checkmark       &  87.0        &  34.5   &   27.1   &       &  97.8  &  76.9 & 64.0 & 44.5 &5.1  \\
 T5    &    E-D      &  \Checkmark         &   86.5        & 36.4    &    27.0  &       & 99.6   & 78.8 & 59.9 & 46.9& 4.6   \\
 Flan-T5      &  E-D        &  \Checkmark         &       67.9    & \textbf{31.8}   & 26.5    &       &  98.6 &   80.8 &   79.7 &  35.3 & 3.9\\
  LlaMa-2      &    D      &  \Checkmark         & 66.9         & 33.4    &  33.0    &       & 94.2  &  77.6  & 62.1 & \textbf{33.2} & 5.2 \\
Mistral     &    D      &     \Checkmark      &  \textbf{52.0}         & 40.8    &   \textbf{25.4}   &       &  \textbf{80.3}  &   76.0  & \textbf{42.3} & 35.0 &  \textbf{3.8} \\
Qwen2.5     &    D      &    \XSolidBrush       &   63.1     &  38.4   &    25.8  &       &  85.0  &  79.3   & 45.1 & 36.0 & 3.9\\
\hline
\end{tabular}}
\end{adjustbox}
\caption{The error rate (\%) of 2-hop game move generation.}
\label{exp1}
\end{table*}

%% file: tables/align.tex
\begin{table}[!ht]
\setlength{\abovecaptionskip}{5pt} 
\setlength{\belowcaptionskip}{0pt}
\centering
 \begin{adjustbox}{max width=0.48\textwidth}
    \setlength{\tabcolsep}{1mm}{
\begin{tabular}{p{1.3cm}p{1.3cm}P{1.2cm}P{1.2cm}cP{1.2cm}P{1.2cm}}
\hline
\multirow{ 2}{*}{Src.} & \multirow{ 2}{*}{Trg.} & \multicolumn{2}{c}{\textbf{Supervised}} & &  \multicolumn{2}{c}{\textbf{Unsupervised}} \\ 
 & & CHAM. & SYN. & & CHAM. & SYN.  \\  
\hline
GPT-2 & Bart &  81.4 & \textbf{93.1} &   & 80.3 & 91.3
\\
GPT-2 & T5& 83.0 & 85.0&  & 76.4 & 80.1 \\
Bart & T5 & 69.2 & 84.5 &  & 85.2  & 81.1\\
GPT-2 & Mistral & \textbf{90.3} & 77.2 &  & 80.3 & 82.6\\
Bart & Mistral &  88.0 & 79.1 &  & \textbf{96.1}  & \textbf{97.2}  \\
LlaMa-2 & Mistral & 80.1  & 74.2 & & 76.2 & 72.6  \\
Qwen2.5 & LlaMa-2 & 84.2 & 80.1 & & 81.3 & 84.9 \\
\hline
\end{tabular}}
\end{adjustbox}
\caption{Representation alignment cosine similarity (\%) results. Src. and Trg. represent source and target space. CHAM., SYN represent CHAMPIONSHIP and SYNTHETIC dataset.}

\label{exp2}
\end{table}

%% file: tables/datastatistics.tex
\begin{table}[]
\centering
 \begin{adjustbox}{max width=0.8\textwidth}
    \setlength{\tabcolsep}{1mm}{
\begin{tabular}{ccc}
\hline
& \textbf{CHAMPIONSHIP} & \textbf{SYNTHETIC}\\
\hline
Num. of Games & 132,588 & 23,796,010 \\
Avg. length & 59.8 $\pm$ 1.5 & 60.0 $\pm$ 0.8 \\
Min. length & 4 & 9  \\ 
Full length portion(\%)& 95.0 & 99.1\\

\hline

\hline
\end{tabular}}
\end{adjustbox}
\caption{Dataset statistics of the two Othello datasets.}
\label{tab:datastatistics}
\end{table}

%% file: iclr2025_conference.bbl
\begin{thebibliography}{37}
\providecommand{\natexlab}[1]{#1}
\providecommand{\url}[1]{\texttt{#1}}
\expandafter\ifx\csname urlstyle\endcsname\relax
  \providecommand{\doi}[1]{doi: #1}\else
  \providecommand{\doi}{doi: \begingroup \urlstyle{rm}\Url}\fi

\bibitem[Abdou et~al.(2021)Abdou, Kulmizev, Hershcovich, Frank, Pavlick, and S{\o}gaard]{Abdou2021CanLM}
Mostafa Abdou, Artur Kulmizev, Daniel Hershcovich, Stella Frank, Ellie Pavlick, and Anders S{\o}gaard.
\newblock Can language models encode perceptual structure without grounding? a case study in color.
\newblock In \emph{Conference on Computational Natural Language Learning}, 2021.
\newblock URL \url{https://api.semanticscholar.org/CorpusID:237491991}.

\bibitem[Barrett et~al.(2019)Barrett, Kementchedjhieva, Elazar, Elliott, and S{\o}gaard]{barrett-etal-2019-adversarial}
Maria Barrett, Yova Kementchedjhieva, Yanai Elazar, Desmond Elliott, and Anders S{\o}gaard.
\newblock Adversarial removal of demographic attributes revisited.
\newblock In Kentaro Inui, Jing Jiang, Vincent Ng, and Xiaojun Wan (eds.), \emph{Proceedings of the 2019 Conference on Empirical Methods in Natural Language Processing and the 9th International Joint Conference on Natural Language Processing (EMNLP-IJCNLP)}, pp.\  6330--6335, Hong Kong, China, November 2019. Association for Computational Linguistics.
\newblock \doi{10.18653/v1/D19-1662}.
\newblock URL \url{https://aclanthology.org/D19-1662}.

\bibitem[Bender \& Koller(2020)Bender and Koller]{bender-koller-2020-climbing}
Emily~M. Bender and Alexander Koller.
\newblock Climbing towards {NLU}: {On} meaning, form, and understanding in the age of data.
\newblock In \emph{Proceedings of the 58th Annual Meeting of the Association for Computational Linguistics}, pp.\  5185--5198, Online, July 2020. Association for Computational Linguistics.
\newblock \doi{10.18653/v1/2020.acl-main.463}.
\newblock URL \url{https://aclanthology.org/2020.acl-main.463}.

\bibitem[Chang et~al.(2018)Chang, Chen, Lin, and Nair]{Chang2018TheBW}
Naiyuan Chang, Chih-Hung Chen, Shun-Shii Lin, and Surag Nair.
\newblock The big win strategy on multi-value network: An improvement over alphazero approach for 6x6 othello.
\newblock \emph{Proceedings of the 2018 International Conference on Machine Learning and Machine Intelligence}, 2018.
\newblock URL \url{https://api.semanticscholar.org/CorpusID:54461634}.

\bibitem[Chung et~al.(2022)Chung, Hou, Longpre, Zoph, Tay, Fedus, Li, Wang, Dehghani, Brahma, Webson, Gu, Dai, Suzgun, Chen, Chowdhery, Valter, Narang, Mishra, Yu, Zhao, Huang, Dai, Yu, Petrov, hsin Chi, Dean, Devlin, Roberts, Zhou, Le, and Wei]{Chung2022ScalingIL}
Hyung~Won Chung, Le~Hou, S.~Longpre, Barret Zoph, Yi~Tay, William Fedus, Eric Li, Xuezhi Wang, Mostafa Dehghani, Siddhartha Brahma, Albert Webson, Shixiang~Shane Gu, Zhuyun Dai, Mirac Suzgun, Xinyun Chen, Aakanksha Chowdhery, Dasha Valter, Sharan Narang, Gaurav Mishra, Adams~Wei Yu, Vincent Zhao, Yanping Huang, Andrew~M. Dai, Hongkun Yu, Slav Petrov, Ed~Huai hsin Chi, Jeff Dean, Jacob Devlin, Adam Roberts, Denny Zhou, Quoc~V. Le, and Jason Wei.
\newblock Scaling instruction-finetuned language models.
\newblock \emph{ArXiv}, abs/2210.11416, 2022.
\newblock URL \url{https://api.semanticscholar.org/CorpusID:253018554}.

\bibitem[Conneau et~al.(2018)Conneau, Lample, Ranzato, Denoyer, and J{\'e}gou]{conneau2017word}
Alexis Conneau, Guillaume Lample, Marc'Aurelio Ranzato, Ludovic Denoyer, and Herv{\'e} J{\'e}gou.
\newblock Word translation without parallel data.
\newblock \emph{The Sixth International Conference on Learning Representations}, 2018.

\bibitem[Gower \& Dijksterhuis(2004)Gower and Dijksterhuis]{gower2004procrustes}
John~C Gower and Garmt~B Dijksterhuis.
\newblock \emph{Procrustes problems}, volume~30.
\newblock OUP Oxford, 2004.

\bibitem[Hao et~al.(2023)Hao, Gu, Ma, Hong, Wang, Wang, and Hu]{Hao2023ReasoningWL}
Shibo Hao, Yi~Gu, Haodi Ma, Joshua~Jiahua Hong, Zhen Wang, Daisy~Zhe Wang, and Zhiting Hu.
\newblock Reasoning with language model is planning with world model.
\newblock \emph{ArXiv}, abs/2305.14992, 2023.
\newblock URL \url{https://api.semanticscholar.org/CorpusID:258865812}.

\bibitem[Harnad(1990)]{Harnad1990-HARTSG}
Stevan Harnad.
\newblock The symbol grounding problem.
\newblock \emph{Physica D}, 42:\penalty0 335--346, 1990.

\bibitem[Hazineh et~al.(2023)Hazineh, Zhang, and Chiu]{Hazineh2023LinearLW}
Dean~S. Hazineh, Zechen Zhang, and Jeffery Chiu.
\newblock Linear latent world models in simple transformers: A case study on othello-gpt.
\newblock \emph{ArXiv}, abs/2310.07582, 2023.
\newblock URL \url{https://api.semanticscholar.org/CorpusID:263834692}.

\bibitem[Hua et~al.(2024)Hua, Yun, and Pavlick]{Hua2024mOthelloWD}
Tianze Hua, Tian Yun, and Ellie Pavlick.
\newblock mothello: When do cross-lingual representation alignment and cross-lingual transfer emerge in multilingual models?
\newblock \emph{ArXiv}, abs/2404.12444, 2024.
\newblock URL \url{https://api.semanticscholar.org/CorpusID:269282665}.

\bibitem[Huh et~al.(2024)Huh, Cheung, Wang, and Isola]{Huh2024ThePR}
Minyoung Huh, Brian Cheung, Tongzhou Wang, and Phillip Isola.
\newblock The platonic representation hypothesis.
\newblock 2024.
\newblock URL \url{https://api.semanticscholar.org/CorpusID:269757765}.

\bibitem[Hui et~al.(2024)Hui, Yang, Cui, Yang, Liu, Zhang, Liu, Zhang, Yu, Dang, Yang, Men, Huang, Quan, Ren, Ren, Zhou, and Lin]{Hui2024Qwen25CoderTR}
Binyuan Hui, Jian Yang, Zeyu Cui, Jiaxi Yang, Dayiheng Liu, Lei Zhang, Tianyu Liu, Jiajun Zhang, Bowen Yu, Kai Dang, An~Yang, Rui Men, Fei Huang, Shanghaoran Quan, Xingzhang Ren, Xuancheng Ren, Jingren Zhou, and Junyang Lin.
\newblock Qwen2.5-coder technical report.
\newblock \emph{ArXiv}, abs/2409.12186, 2024.
\newblock URL \url{https://api.semanticscholar.org/CorpusID:272707390}.

\bibitem[Ivanitskiy et~al.(2023)Ivanitskiy, Spies, Rauker, Corlouer, Mathwin, Quirke, Rager, Shah, Valentine, Behn, Inoue, and Fung]{Ivanitskiy2023StructuredWR}
Michael~I. Ivanitskiy, Alex~F Spies, Tilman Rauker, Guillaume Corlouer, Chris Mathwin, Lucia Quirke, Can Rager, Rusheb Shah, Dan Valentine, Cecilia G.~Diniz Behn, Katsumi Inoue, and Samy~Wu Fung.
\newblock Structured world representations in maze-solving transformers.
\newblock \emph{ArXiv}, abs/2312.02566, 2023.
\newblock URL \url{https://api.semanticscholar.org/CorpusID:265659365}.

\bibitem[Jiang et~al.(2023)Jiang, Sablayrolles, Mensch, Bamford, Chaplot, de~Las~Casas, Bressand, Lengyel, Lample, Saulnier, Lavaud, Lachaux, Stock, Scao, Lavril, Wang, Lacroix, and Sayed]{Jiang2023Mistral7}
Albert~Qiaochu Jiang, Alexandre Sablayrolles, Arthur Mensch, Chris Bamford, Devendra~Singh Chaplot, Diego de~Las~Casas, Florian Bressand, Gianna Lengyel, Guillaume Lample, Lucile Saulnier, L'elio~Renard Lavaud, Marie-Anne Lachaux, Pierre Stock, Teven~Le Scao, Thibaut Lavril, Thomas Wang, Timoth{\'e}e Lacroix, and William~El Sayed.
\newblock Mistral 7b.
\newblock \emph{ArXiv}, abs/2310.06825, 2023.
\newblock URL \url{https://api.semanticscholar.org/CorpusID:263830494}.

\bibitem[Joshi et~al.(2024)Joshi, Sharma, and Modi]{Joshi2024CheckersGPTLW}
Abhinav Joshi, Vaibhav Sharma, and Ashutosh Modi.
\newblock Checkersgpt: Learning world models through language modeling.
\newblock In \emph{Annual Meeting of the Association for Computational Linguistics}, 2024.
\newblock URL \url{https://api.semanticscholar.org/CorpusID:272779453}.

\bibitem[Karvonen(2024)]{Karvonen2024EmergentWM}
Adam Karvonen.
\newblock Emergent world models and latent variable estimation in chess-playing language models.
\newblock \emph{ArXiv}, abs/2403.15498, 2024.
\newblock URL \url{https://api.semanticscholar.org/CorpusID:268681535}.

\bibitem[Lample et~al.(2018)Lample, Conneau, Denoyer, and Ranzato]{lample2017unsupervised}
Guillaume Lample, Alexis Conneau, Ludovic Denoyer, and Marc'Aurelio Ranzato.
\newblock Unsupervised machine translation using monolingual corpora only.
\newblock \emph{The Sixth International Conference on Learning Representations}, 2018.

\bibitem[Lewis et~al.(2019)Lewis, Liu, Goyal, Ghazvininejad, rahman Mohamed, Levy, Stoyanov, and Zettlemoyer]{Lewis2019BARTDS}
Mike Lewis, Yinhan Liu, Naman Goyal, Marjan Ghazvininejad, Abdel rahman Mohamed, Omer Levy, Veselin Stoyanov, and Luke Zettlemoyer.
\newblock Bart: Denoising sequence-to-sequence pre-training for natural language generation, translation, and comprehension.
\newblock In \emph{Annual Meeting of the Association for Computational Linguistics}, 2019.
\newblock URL \url{https://api.semanticscholar.org/CorpusID:204960716}.

\bibitem[Li et~al.(2021)Li, Nye, and Andreas]{Li2021ImplicitRO}
Belinda~Z. Li, Maxwell Nye, and Jacob Andreas.
\newblock Implicit representations of meaning in neural language models.
\newblock In \emph{Annual Meeting of the Association for Computational Linguistics}, 2021.
\newblock URL \url{https://api.semanticscholar.org/CorpusID:235294296}.

\bibitem[Li et~al.(2023)Li, Hopkins, Bau, Vi{\'e}gas, Pfister, and Wattenberg]{li2023emergent}
Kenneth Li, Aspen~K Hopkins, David Bau, Fernanda Vi{\'e}gas, Hanspeter Pfister, and Martin Wattenberg.
\newblock Emergent world representations: Exploring a sequence model trained on a synthetic task.
\newblock In \emph{The Eleventh International Conference on Learning Representations}, 2023.
\newblock URL \url{https://openreview.net/forum?id=DeG07_TcZvT}.

\bibitem[Liskowski et~al.(2018)Liskowski, Jaśkowski, and Krawiec]{8276588}
Paweł Liskowski, Wojciech Jaśkowski, and Krzysztof Krawiec.
\newblock Learning to play othello with deep neural networks.
\newblock \emph{IEEE Transactions on Games}, 10\penalty0 (4):\penalty0 354--364, 2018.
\newblock \doi{10.1109/TG.2018.2799997}.

\bibitem[Mikolov et~al.(2013)Mikolov, Sutskever, Chen, Corrado, and Dean]{NIPS2013_9aa42b31}
Tomas Mikolov, Ilya Sutskever, Kai Chen, Greg~S Corrado, and Jeff Dean.
\newblock Distributed representations of words and phrases and their compositionality.
\newblock In C.J. Burges, L.~Bottou, M.~Welling, Z.~Ghahramani, and K.Q. Weinberger (eds.), \emph{Advances in Neural Information Processing Systems}, volume~26. Curran Associates, Inc., 2013.
\newblock URL \url{https://proceedings.neurips.cc/paper_files/paper/2013/file/9aa42b31882ec039965f3c4923ce901b-Paper.pdf}.

\bibitem[Nanda et~al.(2023{\natexlab{a}})Nanda, Lee, and Wattenberg]{Nanda2023EmergentLR}
Neel Nanda, Andrew Lee, and Martin Wattenberg.
\newblock Emergent linear representations in world models of self-supervised sequence models.
\newblock \emph{ArXiv}, abs/2309.00941, 2023{\natexlab{a}}.
\newblock URL \url{https://api.semanticscholar.org/CorpusID:261530966}.

\bibitem[Nanda et~al.(2023{\natexlab{b}})Nanda, Lee, and Wattenberg]{nanda-etal-2023-emergent}
Neel Nanda, Andrew Lee, and Martin Wattenberg.
\newblock Emergent linear representations in world models of self-supervised sequence models.
\newblock In Yonatan Belinkov, Sophie Hao, Jaap Jumelet, Najoung Kim, Arya McCarthy, and Hosein Mohebbi (eds.), \emph{Proceedings of the 6th BlackboxNLP Workshop: Analyzing and Interpreting Neural Networks for NLP}, pp.\  16--30, Singapore, December 2023{\natexlab{b}}. Association for Computational Linguistics.
\newblock \doi{10.18653/v1/2023.blackboxnlp-1.2}.
\newblock URL \url{https://aclanthology.org/2023.blackboxnlp-1.2}.

\bibitem[Noever \& Noever(2022)Noever and Noever]{noever2022word}
Samantha E.~Miller Noever and David Noever.
\newblock Word play for playing othello (reverses), 2022.

\bibitem[Patel \& Pavlick(2022)Patel and Pavlick]{Patel2022MappingLM}
Roma Patel and Ellie Pavlick.
\newblock Mapping language models to grounded conceptual spaces.
\newblock In \emph{International Conference on Learning Representations}, 2022.
\newblock URL \url{https://api.semanticscholar.org/CorpusID:251647156}.

\bibitem[Radford et~al.(2019)Radford, Wu, Child, Luan, Amodei, and Sutskever]{Radford2019LanguageMA}
Alec Radford, Jeff Wu, Rewon Child, David Luan, Dario Amodei, and Ilya Sutskever.
\newblock Language models are unsupervised multitask learners.
\newblock 2019.
\newblock URL \url{https://api.semanticscholar.org/CorpusID:160025533}.

\bibitem[Raffel et~al.(2019)Raffel, Shazeer, Roberts, Lee, Narang, Matena, Zhou, Li, and Liu]{Raffel2019ExploringTL}
Colin Raffel, Noam~M. Shazeer, Adam Roberts, Katherine Lee, Sharan Narang, Michael Matena, Yanqi Zhou, Wei Li, and Peter~J. Liu.
\newblock Exploring the limits of transfer learning with a unified text-to-text transformer.
\newblock \emph{J. Mach. Learn. Res.}, 21:\penalty0 140:1--140:67, 2019.
\newblock URL \url{https://api.semanticscholar.org/CorpusID:204838007}.

\bibitem[S{\o}gaard et~al.(2019)S{\o}gaard, Vuli{\'c}, Ruder, and Faruqui]{4264a46fd9e846e4a704b2d13002e521}
Anders S{\o}gaard, Ivan Vuli{\'c}, Sebastian Ruder, and Manaal Faruqui.
\newblock \emph{Cross-Lingual Word Embeddings}.
\newblock Synthesis Lectures on Human Language Technologies. Morgan \& Claypool Publishers, United States, 2 edition, 2019.
\newblock \doi{10.2200/S00920ED2V01Y201904HLT042}.

\bibitem[Takizawa(2024)]{takizawa2024othello}
Hiroki Takizawa.
\newblock Othello is solved, 2024.

\bibitem[Toshniwal et~al.(2021)Toshniwal, Wiseman, Livescu, and Gimpel]{Toshniwal2021LearningCB}
Shubham Toshniwal, Sam Wiseman, Karen Livescu, and Kevin Gimpel.
\newblock Learning chess blindfolded: Evaluating language models on state tracking.
\newblock \emph{ArXiv}, abs/2102.13249, 2021.
\newblock URL \url{https://api.semanticscholar.org/CorpusID:232069003}.

\bibitem[Touvron et~al.(2023)Touvron, Martin, Stone, Albert, Almahairi, Babaei, Bashlykov, Batra, Bhargava, Bhosale, Bikel, Blecher, Ferrer, Chen, Cucurull, Esiobu, Fernandes, Fu, Fu, Fuller, Gao, Goswami, Goyal, Hartshorn, Hosseini, Hou, Inan, Kardas, Kerkez, Khabsa, Kloumann, Korenev, Koura, Lachaux, Lavril, Lee, Liskovich, Lu, Mao, Martinet, Mihaylov, Mishra, Molybog, Nie, Poulton, Reizenstein, Rungta, Saladi, Schelten, Silva, Smith, Subramanian, Tan, Tang, Taylor, Williams, Kuan, Xu, Yan, Zarov, Zhang, Fan, Kambadur, Narang, Rodriguez, Stojnic, Edunov, and Scialom]{Touvron2023Llama2O}
Hugo Touvron, Louis Martin, Kevin~R. Stone, Peter Albert, Amjad Almahairi, Yasmine Babaei, Nikolay Bashlykov, Soumya Batra, Prajjwal Bhargava, Shruti Bhosale, Daniel~M. Bikel, Lukas Blecher, Cristian~Cant{\'o}n Ferrer, Moya Chen, Guillem Cucurull, David Esiobu, Jude Fernandes, Jeremy Fu, Wenyin Fu, Brian Fuller, Cynthia Gao, Vedanuj Goswami, Naman Goyal, Anthony~S. Hartshorn, Saghar Hosseini, Rui Hou, Hakan Inan, Marcin Kardas, Viktor Kerkez, Madian Khabsa, Isabel~M. Kloumann, A.~V. Korenev, Punit~Singh Koura, Marie-Anne Lachaux, Thibaut Lavril, Jenya Lee, Diana Liskovich, Yinghai Lu, Yuning Mao, Xavier Martinet, Todor Mihaylov, Pushkar Mishra, Igor Molybog, Yixin Nie, Andrew Poulton, Jeremy Reizenstein, Rashi Rungta, Kalyan Saladi, Alan Schelten, Ruan Silva, Eric~Michael Smith, R.~Subramanian, Xia Tan, Binh Tang, Ross Taylor, Adina Williams, Jian~Xiang Kuan, Puxin Xu, Zhengxu Yan, Iliyan Zarov, Yuchen Zhang, Angela Fan, Melanie Kambadur, Sharan Narang, Aurelien Rodriguez, Robert Stojnic, Sergey Edunov, and
  Thomas Scialom.
\newblock Llama 2: Open foundation and fine-tuned chat models.
\newblock \emph{ArXiv}, abs/2307.09288, 2023.
\newblock URL \url{https://api.semanticscholar.org/CorpusID:259950998}.

\bibitem[van~der Ree \& Wiering(2013)van~der Ree and Wiering]{Ree2013ReinforcementLI}
Michiel van~der Ree and Marco~A Wiering.
\newblock Reinforcement learning in the game of othello: Learning against a fixed opponent and learning from self-play.
\newblock \emph{2013 IEEE Symposium on Adaptive Dynamic Programming and Reinforcement Learning (ADPRL)}, pp.\  108--115, 2013.
\newblock URL \url{https://api.semanticscholar.org/CorpusID:1695896}.

\bibitem[Wang et~al.(2024)Wang, Todd, Xiao, Yuan, Cot'e, Clark, and Jansen]{Wang2024CanLM}
Ruoyao Wang, Graham Todd, Ziang Xiao, Xingdi Yuan, Marc-Alexandre Cot'e, Peter Clark, and Peter Jansen.
\newblock Can language models serve as text-based world simulators?
\newblock 2024.
\newblock URL \url{https://api.semanticscholar.org/CorpusID:270371867}.

\bibitem[Xiang et~al.(2023)Xiang, Tao, Gu, Shu, Wang, Yang, and Hu]{Xiang2023LanguageMM}
Jiannan Xiang, Tianhua Tao, Yi~Gu, Tianmin Shu, Zirui Wang, Zichao Yang, and Zhiting Hu.
\newblock Language models meet world models: Embodied experiences enhance language models.
\newblock \emph{ArXiv}, abs/2305.10626, 2023.
\newblock URL \url{https://api.semanticscholar.org/CorpusID:258762577}.

\bibitem[Yun et~al.(2023)Yun, Zeng, Handa, Thapliyal, Pang, Pavlick, and Sun]{Yun2023EmergenceOA}
Tian Yun, Zilai Zeng, Kunal Handa, Ashish~V. Thapliyal, Bo~Pang, Ellie Pavlick, and Chen Sun.
\newblock Emergence of abstract state representations in embodied sequence modeling.
\newblock In \emph{Conference on Empirical Methods in Natural Language Processing}, 2023.
\newblock URL \url{https://api.semanticscholar.org/CorpusID:265034007}.

\end{thebibliography}
